\documentclass[9.5pt,journal,compsoc]{IEEEtran}

\ifCLASSOPTIONcompsoc
  \usepackage[nocompress]{cite}
\else
  \usepackage{cite}
\fi

\usepackage{amsmath}
\usepackage{color}
\usepackage{array}
\usepackage{breqn}
\usepackage{{inputenc}}
\usepackage{balance}
\usepackage{multirow,tabularx}
\usepackage{booktabs,longtable}
\usepackage{hhline}
\usepackage[flushleft]{threeparttable}
\usepackage{algorithm}
\usepackage{algorithmic}
\usepackage{epsfig}
\usepackage{graphicx}
\usepackage{epstopdf}
\usepackage{thmtools}
\usepackage{enumitem}
\usepackage{verbatim}
\usepackage{amsfonts}
\usepackage[colorlinks=false,
            linkcolor=red,
            anchorcolor=blue,
            citecolor=green
            ]{hyperref}
\usepackage{xcolor}
\usepackage{subcaption}

\newcommand{\Mat}[1]{\textbf{#1}}

\newcommand{\Space}[1]{\mathbb{#1}}
\newcommand{\Set}[1]{\mathcal{#1}}

\newcommand{\ie}{\emph{i.e., }}
\newcommand{\eg}{\emph{e.g., }}

\newcommand{\wrt}{\emph{w.r.t. }}
\newcommand{\cf}{\emph{cf. }}

\newcommand{\wx}[1]{{\color{black}{#1}}}
\newcommand{\wxx}[1]{{\color{black}{#1}}}

\begin{document}

\title{Reinforced Causal Explainer for Graph Neural Networks}

\author{Xiang Wang,
        Yingxin Wu,
        An Zhang,
        Fuli Feng,
        Xiangnan He,
        Tat-Seng Chua
\IEEEcompsocitemizethanks{
  \IEEEcompsocthanksitem X. Wang, Y. Wu, F. Feng, X. He are with University of Science and Technology of China; CCCD Key Lab of Ministry of Culture and Tourism. E-mail: xiangwang1223@gmail.com, wuyxin@mail.ustc.edu.cn, fulifeng93@gmail.com, xiangnanhe@gmail.com.
  \protect\\

  \IEEEcompsocthanksitem A. Zhang and T. Chua are with National University of Singapore. E-mail: an\_zhang@nus.edu.sg,  dcscts@nus.edu.sg. A. Zhang is the corresponding author.}
}

%
%

\markboth{IEEE Transactions on Pattern Analysis and Machine Intelligence}%
{Wang \MakeLowercase{\textit{et al.}}: Reinforced Causal Explainer for Graph Neural Networks}
%



\IEEEtitleabstractindextext{%
\begin{abstract}
  Explainability is crucial for probing graph neural networks (GNNs), answering questions like ``Why the GNN model makes a certain prediction?''.
  Feature attribution is a prevalent technique of highlighting the explanatory subgraph in the input graph, which plausibly leads the GNN model to make its prediction.
  Various attribution methods have been proposed to exploit gradient-like or attention scores as the attributions of edges, then select the salient edges with top attribution scores as the explanation.
  However, most of these works make an untenable assumption --- the selected edges are linearly independent --- thus leaving the dependencies among edges largely unexplored, especially their coalition effect.
  We demonstrate unambiguous drawbacks of this assumption --- making the explanatory subgraph unfaithful and verbose.
  To address this challenge, we propose a reinforcement learning agent, \emph{Reinforced Causal Explainer} (RC-Explainer).
  It frames the explanation task as a sequential decision process --- an explanatory subgraph is successively constructed by adding a salient edge to connect the previously selected subgraph.
  Technically, its policy network predicts the action of edge addition, and gets a reward that quantifies the action's causal effect on the prediction.
  Such reward accounts for the dependency of the newly-added edge and the previously-added edges, thus reflecting whether they collaborate together and form a coalition to pursue better explanations.
  It is trained via policy gradient to optimize the reward stream of edge sequences.
  As such, RC-Explainer is able to generate faithful and concise explanations, and has a better generalization power to unseen graphs.
  When explaining different GNNs on three graph classification datasets, RC-Explainer achieves better or comparable performance to state-of-the-art approaches \wrt two quantitative metrics: predictive accuracy, contrastivity, and safely passes sanity checks and visual inspections.
  \wxx{Codes and datasets are available at \url{https://github.com/xiangwang1223/reinforced_causal_explainer}.}
\end{abstract}

}

\maketitle

\IEEEdisplaynontitleabstractindextext

\IEEEpeerreviewmaketitle

\IEEEraisesectionheading{\section{Introduction}\label{sec:introduction}}



\IEEEPARstart{G}raph neural networks (GNNs)~\cite{GraphSage,Benchmarking} have exhibited impressive performance in a variety of tasks, where the data are graph-structured.
Their success comes mainly from the powerful representation learning, which incorporates graph structure in an end-to-end fashion.
Alongside performance, explainability becomes central to the practical impact of GNNs, especially in real-world applications on fairness, security, and robustness \cite{XGNN,SubgraphX,ShadeWatcher}.
Aiming to answer questions like ``Why the GNN model made a certain prediction?'', we focus on post-hoc \cite{rudin2019stop}, local \cite{LIME}, model-agnostic \cite{DeepLIFT} explainability --- that is, considering the target GNN model as a black-box (\ie post-hoc), an explainer interprets its predictions of individual instances (\ie local), which is applicable to any GNNs (\ie model-agnostic).
Towards this end, a prevalent paradigm is feature attribution and selection \cite{Grad-CAM,IG,DBLP:conf/icml/ChattopadhyayMS19,Towardsbetter}.
Typically, given an input graph, it distributes the model's outcome prediction to the input features (\ie edges), and then selects the salient substructure (\ie the subset of edges) as an explanatory subgraph.
Such an explanatory subgraph is expected to provide insight into the model workings.


Towards high-quality feature attribution, it is essential to uncover the relationships between the input features and outcome predictions underlying the GNN model.
Most existing explainers reveal the relationships via
(1) gradient-like signals \wrt edges \cite{SA-Graph,Grad-CAM-Graph,GNN-LRP}, which are obtained by backpropagating the model outcome to the graph structure;
(2) masks or attention scores of edges \cite{GNNExplainer,PGExplainer,ReFine}, which are derived from the masking functions or attention networks to approximate the target prediction via the fractional (masked or attentive) graph;
or (3) prediction changes on perturbed edges \cite{PGM-Explainer,CXPlain,SubgraphX}, which are fetched by perturbing the graph structure, such as leaving subgraphs out and auditing the outcome change \cite{PGM-Explainer} or computing the Shapley values \cite{SubgraphX}.
Subsequently, the subset of edges with top attributions composes an explanatory subgraph most influential to the model's decision. 

\begin{figure}[t]
    \centering
	\includegraphics[width=0.96\columnwidth]{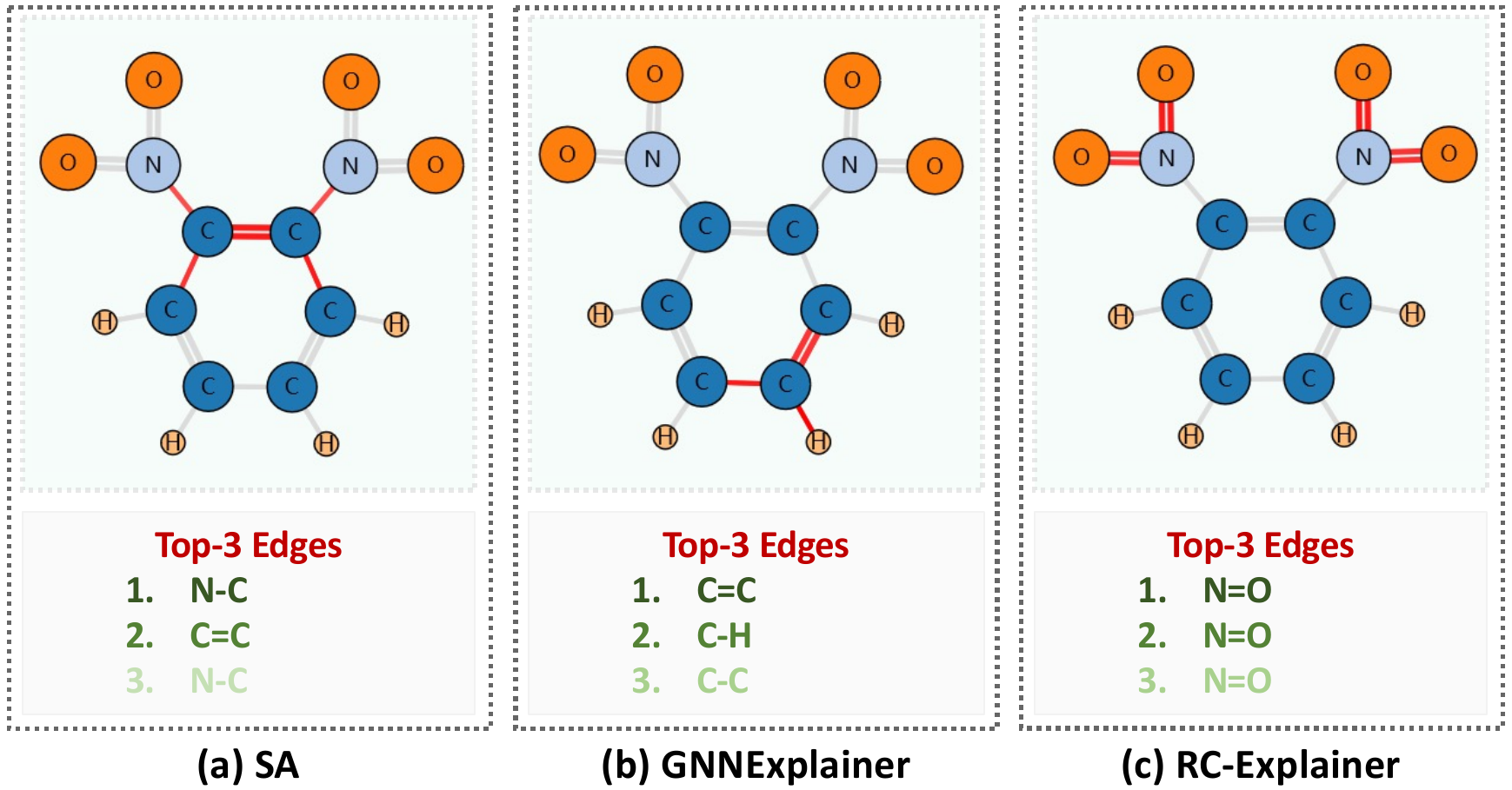}
	\vspace{-5pt}
	\caption{A real example of explaining the mutagenicity classification of a molecule graph. (a-c) show explanations of SA, GNNExplainer, and RC-Explainer, respectively, where the important edges are highlighted with red color and the top-3 edges are listed. Best viewed in color.}
	\label{fig:intro}
	\vspace{-15pt}
\end{figure}

Nonetheless, we argue that these explainers are prone to generating suboptimal explanations since two key factors remain largely unexplored:
\begin{itemize}
    \item Causal effects of edges. It is crucial to specify the edges that are the plausible causation of the model outcome, rather than the edges irrelevant or spuriously correlated with the outcome \cite{moraffah2020causal,DBLP:conf/wsdm/Pearl18}. However, as the explainers using gradient- and attention-like scores typically approach the input-outcome relationships from an associational standpoint, they hardly distinguish causal from noncausal effects of edges.
    Take Figure \ref{fig:intro} as a running example, where SA \cite{SA-Graph} and GNNExplainer \cite{GNNExplainer} explain why the GIN model \cite{DBLP:conf/iclr/XuHLJ19} predicts the molecule graph as \emph{mutagenic}.
    As the nitrogen-carbon (\emph{N-C}) bond often connects with the nitro group (\emph{NO$_{2}$}), it is spuriously correlated with the \emph{mutagenic} property, thus ranked top by SA.
    Feeding such spurious edges solely into the model, however, hardly recovers the target prediction, thus unfaithfully reflect the model's decision.
    
    \item Dependencies among edges. Most explainers ignore the edge dependencies when probing the edge attributions and constructing the explanatory subgraph. One key reason is that they draw attributions of edges independently \cite{DBLP:conf/nips/LundbergL17,PGM-Explainer,DBLP:journals/corr/abs-2006-01272}.
    In fact, edges usually collaborate and cooperate with other edges to approach the decision boundary instead. Such highly-dependent edges form a coalition that can frame a prototype memorized in the model to make decisions.
    Considering SA's explanation, compared to the individual \emph{N-C} bond, its combination with the carbon-carbon (\emph{C=C}) bond brings no unique information about the model prediction, as only a marginal improvement on predictive accuracy is achieved\footnote{When removing the counterparts, the single \emph{N-C} bond and its combination with the \emph{C=C} bond are predicted as mutagenic with probability of 0.31 and 0.35, respectively.}. In contrast, two nitrogen-oxygen (\emph{N=O}) bonds \wx{form a nitro group} (\emph{NO$_2$}), which is a typical coalition responsible for the mutagenic property \cite{XGNN} \wx{and purses a higher influence on the prediction than individuals}\footnote{The probability of being mutagenic increases from 0.72 (feeding the first \emph{N=O} bond solely into GIN) to 0.95 (feeding the first two \emph{N=O} bonds together into GIN).}. Clearly, the \emph{N=O} bonds within \emph{NO$_2$} ought to be better post-hoc explanations.
\end{itemize}

In this work, we explore the edges' causal effects and dependencies, in order to generate explanations that are faithful to the model's decision-making process and consistent to human cognition.
Indeed, it is challenging but can be solved after we introduce the screening strategy \cite{aliferis2010local} equipped with causality \cite{pearl2009causality}.
Specifically, the screening strategy frames the explanation task as sequentially adding edges --- that is, it starts from an empty set as the explanatory subgraph, and incrementally adds edges into the subgraph, one edge at a time step.
During the screening, causality gifts us the manner to assess the dependencies of the previously added edges and the edge candidate being selected, answering the ``What would happen to the prediction, if we add this edge into the GNN’s input?'' question.
The basis is doing causal attribution on the edge candidate as its individual causal effect (ICE) \cite{pearl2009causality,pearl2000causality}, given the previous selections.
Formally, it compares the model outcomes under treatment (\ie the GNN takes the coalition of the edge and the previous selection as input) and control (\ie the GNN tasks the previous selection only).
Positive causal attribution indicates the edge coalition offers unique information strongly relevant to the prediction; otherwise, the edge is redundant or irrelevant.

Towards this end, we propose a reinforcement learning (RL) agent, \textbf{Reinforced Causal Explainer} (RC-Explainer), to achieve the causal screening strategy.
The insight is that the RL agent with the stochastic policy can determine automatically where to search, given the uncertainty information of the learned policy, which can be updated promptly by the stream of reward signals.
Technically, RC-Explainer uses a simple neural network as the policy network to learn the probability of edge candidates being selected, and then select one potential edge as the action, at each step.
Such a sequence of edges forms the policy, and gets rewards that consist of the causal attributions of each compositional edge and subgraphs.
As such, we can to exhibit the dependencies of explained edges, and highlight the influence of edge coalition.
We resort to policy gradient to optimize the policy network.
With a global understanding of the GNN, our RC-Explainer is able to offer model-level explanations for each graph instance, and generalize to unseen graphs.
Experiments on three datasets showcase the effectiveness of our explanations, which are more consistent, concise, and faithful to the predictions compared with existing methods.

Contributions of this work are summarized as follows:
\begin{itemize}
    \item \wx{We emphasize the importance of edges' causal effects and dependencies to reveal the edge attributions and build the explanatory subgraph, so as to interpret the GNN predictions faithfully and concisely.}
    \item We frame the explanation as a sequential decision process and develop an RL agent, RC-Explainer, to take the cause-effect look at the edge dependencies.
    \item We conduct extensive experiments on three datasets, showing the effectiveness of our RC-Explainer \wrt predictive accuracy, contrastivity, sanity checks, and visual inspections.
\end{itemize}


\section{Related Work}
In the literature of explainers, there are many dichotomies approaching explanations --- between post-hoc and intrinsic \cite{rudin2019stop,DIR,IGV} (\ie the target model is explained post hoc by an additional explainer method or is inherently interpretable), between local and global \cite{LIME} (\ie the explainer offers explanations for data instances individually or holistically), between model-agnostic and model-specific \cite{DeepLIFT} (\ie the explainer is comparable across model types or customized for a specific model).
We focus on post-hoc, local, and model-agnostic explainability in this work.

\subsection{Explainability in Non-Graph Neural Networks}
As a prevalent technique, feature attribution \cite{Grad-CAM,IG,DBLP:conf/icml/ChattopadhyayMS19,Towardsbetter} has shown great potential to generate post-hoc explanations for neural networks, especially convolutional neural networks (CNNs).
In general, the existing attribution methods roughly fall into three groups:
\begin{itemize}
    \item One research line \cite{gradients,Grad-CAM,LRP,IG,DeepLIFT} decomposes the model outcome to the input features via backpropagation, such that the gradient-like signals are viewed as the importance of input features.
    For example, Gradient~\cite{gradients} uses pure gradients \wrt input features.
    Grad-CAM \cite{Grad-CAM} additionally leverages the layer-wise context to improve gradients.

    \item Another research line \cite{L2X,VIBI} introduces the trainable attention or masking networks and generates attention scores on input features.
    The networks are trained to approximate the decision boundary of the model via the attentive or masked features.
    For example, L2X \cite{L2X} learns to generate feature masks, with the aim of maximizing the mutual information between the masked features and the target predictions.

    \item Some works \cite{LIME,CXPlain} perform the input perturbations and monitor the changes on model behaviors (\eg prediction, loss), so as to reveal the input-outcome relationships.
    The basic idea is that the model outcomes are highly likely to significantly change if essential features are occluded.
    For example, CXPlain \cite{CXPlain} learns the marginal effect of a feature by occluding it.
\end{itemize}


\subsection{Explainability in Graph Neural Networks}
Compared to the extensive studies in CNNs, explainability in GNNs is less explored and remains a challenging open problem.
Inspired by the methods devised for CNNs, recent works explaining GNNs by:
\begin{itemize}
    \item Gradient-like signals \wrt the graph structure \cite{SA-Graph,Grad-CAM-Graph,GNN-LRP}: For example, SA \cite{SA-Graph} adopts gradients of GNN's loss \wrt adjacency matrix as edge scores, while Grad-CAM \cite{Grad-CAM-Graph} is extended on GNNs.
    
    \item Masks or attention scores of structural features \cite{GNNExplainer,PGExplainer,ReFine}: The basic idea is to maximize the mutual information between the fractional (attentive) graph and the target prediction.
    For instance, GNNExplainer \cite{GNNExplainer} customizes masks on the adjacency matrix for each graph independently.
    Later, PGExplainer \cite{PGExplainer} trains a neural network to generate masks for multiple graphs collectively.
    More recently, ReFine \cite{ReFine} first pre-trains an attention network over class-wise graphs to latch on the global explanation view, and then fine-tunes the local explanations for individual graphs.

    \item Prediction changes on structure perturbations \cite{CXPlain,PGM-Explainer,GraphLIME,SubgraphX}:
    To get the node attributions, PGM-Explainer \cite{PGM-Explainer} applies random perturbations on nodes and learns a Bayesian network upon the perturbation-prediction data to identify significant nodes.
    More recently, SubgraphX \cite{SubgraphX} employs the Monte Carlo tree search algorithm to explore different subgraphs and uses Shapley values to measure each subgraph's importance.
\end{itemize}
Note that XGNN \cite{XGNN} focuses on model-level explanations, rather than local explanations for individual predictions.
Moreover, it fails to preserve the local fidelity of individual graphs, as its explanation may not be a substructure existing in the input graph.
In contrast, our RC-Explainer explains each graph with a global view of the GNN model, which can preserve the local fidelity.

As suggested in the previous work \cite{PGM-Explainer}, most of these works assume that the features are attributed independently, but ignore their dependencies, especially the coalition effect.
Our work differs from them --- we reformulate the explanation generation as a sequential decision process, accounting for the edge relationships and causal effects at each step, towards more faithful and concise explanations.

\section{Preliminaries}

We first summarize the background of GNNs, and then describe the task of generating local, post-hoc explanations.

\subsection{Background of GNNs}\label{sec:background-of-gnns}
Let represent a graph-structured data instance as $\Set{G}=\{e|e\in\Set{E}\}$, where one edge $e=(v,u)\in\Set{E}$ involves two nodes $v$ and $u\in\Set{V}$, to highlight the structural features (\ie the presence of an edge and its endpoints).
Typically, each node $v$ is assigned with a $d$-dimensional feature $\Mat{x}_{v}\in\Space{R}^{d}$.

Various GNNs~\cite{NeuralMessagePassing,GraphSage,GAT,Benchmarking} have been proposed to incorporate such structural features into the representation learning in an end-to-end fashion, so as to facilitate the downstream prediction tasks.
\wx{Following the supervised learning paradigm, we can systematize the GNN model $f$ as a combination of two modules $f_{2}\circ  f_{1}$, where $f_{1}$ is the encoder to generate representations and $f_{2}$ is the predictor to output the predictions.
Clearly, representation learning in the encoder is at the core of GNNs, which typically involves two crucial components:}
\begin{itemize}
    \item Representation aggregation, which \wx{distills} the vectorized information from the adjacent nodes to update the representations of ego nodes recursively:
    \begin{align}\label{equ:gnn-aggregation-part}
        \Mat{a}^{(l)}_{v}&=\text{AGGREGATE}^{(l)}(\{\Mat{z}^{(l-1)}_{u}|u\in\Set{N}_{v}\}),\nonumber\\
        \Mat{z}^{(l)}_{v}&=\text{UPDATE}^{(l)}(\Mat{z}^{(l-1)}_{v},\Mat{a}^{(l)}_{v}),
    \end{align}
    where $\Mat{a}^{(l)}_{v}$ is the aggregation of information propagated from $v$'s neighbors $\Set{N}_{v}=\{u|(v,u)\in \Set{E}\}$; $\Mat{z}^{(0)}_{v}$ is initialized with $\Mat{x}_{v}$, and $\Mat{z}^{(l)}_{v}$ is $v$'s representation after $l$ layers; $\text{AGGREGATE}^{(l)}(\cdot)$ and $\text{UPDATE}^{(l)}(\cdot)$ denote the aggregation and update functions, respectively.
    
    \item Representation readout, which finalizes the representations of node $v$ or graph $\Set{G}$ for the prediction tasks:
    \begin{align}
        \Mat{z}_{v}&=\text{EDGE-READOUT}(\{\Mat{z}^{(l)}_{v}|l=[0,\cdots,L]\}),\label{equ:gnn-node-representation}\\
        \Mat{z}_{\Set{G}}&=\text{GRAPH-READOUT}(\{\Mat{z}_{v}|v\in \Set{V}\}),\label{equ:gnn-graph-representation}
    \end{align}
    where $\Mat{z}_{v}$ is used for the node-level tasks (\eg node classification, link prediction), while $\Mat{z}_{\Set{G}}$ is used for the graph-level tasks (\eg graph classification, graph regression, graph matching); $\text{EDGE-READOUT}(\cdot)$ and $\text{GRAPH-READOUT}(\cdot)$ represent readout functions of nodes and graphs, respectively.
\end{itemize}
Having obtained the powerful representations, \wx{the predictor performs predictions}, such as hiring inner product over two node representations to do link prediction, or employing neural networks on a single node representation to perform node classification.

Without loss of generality, we consider the graph classification problem in this work.
Let $f:\Space{G}\rightarrow \{1,\cdots,C\}$ be a trained GNN model, which classifies a graph instance $\Set{G}\in\Space{G}$ into $C$ classes:
\begin{gather}
    \hat{y}_{c}=f(\Set{G})=f_{2}\circ f_{1}(\Set{G}),
\end{gather}
where $\hat{y}_{c}$ is the predicted class being explained, which is assigned with the largest probability $p_{\theta}(\hat{y}_{c}|\Set{G})$; $f$ is parameterized by $\theta$ \wx{including parameters of encoder and predictor.}


\subsection{Task Description}

Explainability of GNNs aims to answer the questions like ``Given a graph instance $\Set{G}$ of interest, what determines the GNN model $f$ to making a certain output $\hat{y}_{c}$?''.
A prevalent technique to offer local, post-hoc, and model-agnostic explanations is feature attribution~\cite{Grad-CAM,IG,DBLP:conf/icml/ChattopadhyayMS19,Towardsbetter}, which decomposes the prediction to the input features.
As such, each input feature of graph is associated with an attribution score to indicate how much it contributes to the prediction.

Formally, the task of an explainer is to derive the top $K$ important edges and construct a faithful explanatory subgraph $\Set{G}^{*}_{K}=\{e^{*}_{1},\cdots,e^{*}_{K}\}\subseteq\Set{G}$, such that $\Set{G}^{*}_{K}$ offers evidence supporting $f$'s prediction $\hat{y}_{c}$.
Wherein, $e^{*}_{k}$ is the edge ranked at the $k$-th position.
In this work, we follow the prior studies~\cite{GNNExplainer,PGExplainer,Grad-CAM-Graph} and focus mainly on the structural features (\ie the existence of an edge and its endpoints), leaving the identification of salient content features (\ie node features) in future work.
\section{Methodology}

\begin{figure*}[t]
    \centering
	\includegraphics[width=0.96\textwidth]{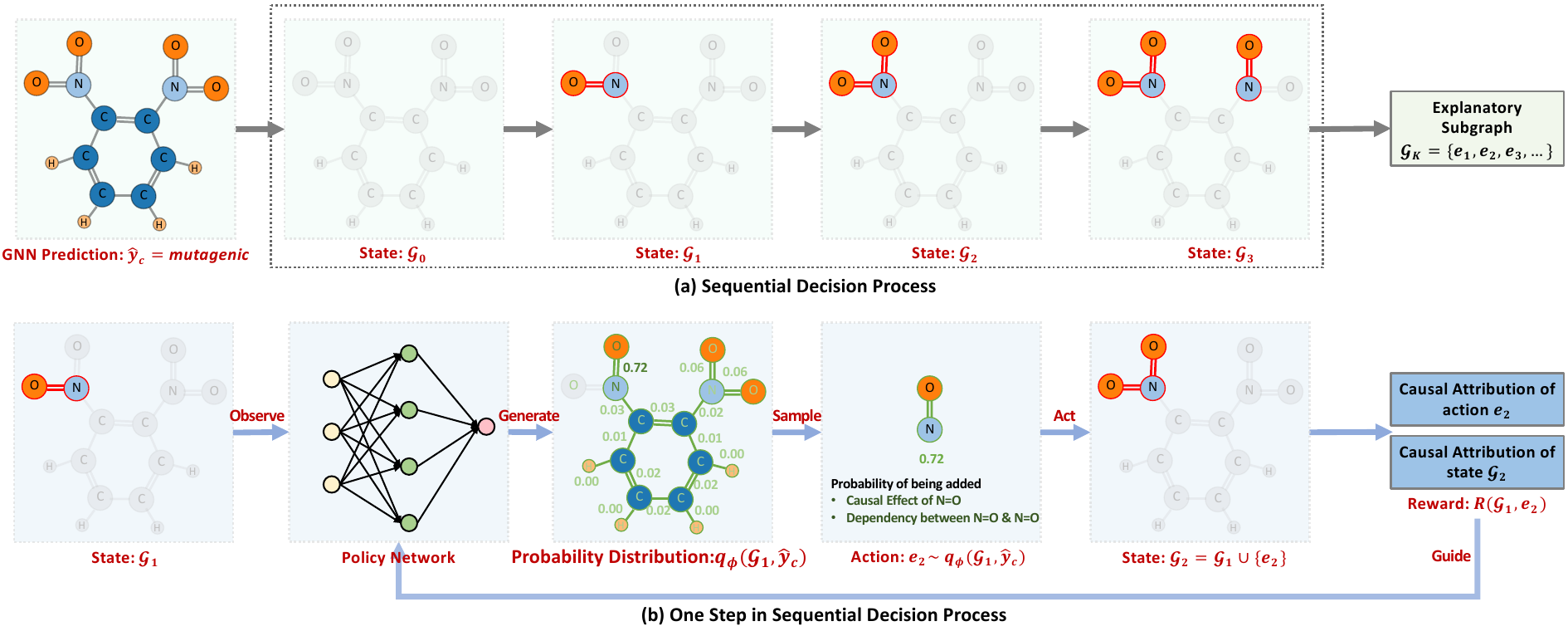}
	\vspace{-5pt}
	\caption{Illustration of the proposed RC-Explainer framework. (a) Sequential decision process to generate the explanatory subgraph; (b) One step from the state $\Set{G}_{1}$ to the state $\Set{G}_{2}$. Best viewed in color.}
	\label{fig:framework}
	\vspace{-10pt}
\end{figure*}

\wx{Towards generating an explanatory subgraph, we first frame the attribution of the holistic subgraph from the causality standpoint and emphasize its limitations.
We then propose the attribution of the edge sequence (termed causal screening), which sequentially selects one edge by measuring its causal effect on the target prediction and its dependencies on the previously-selected edges.
To efficiently achieve the idea of causal screening, we devise a reinforcement learning agent, Reinforced Causal Explainer (RC-Explainer).
}


    

\subsection{Causal Attribution of A Holistic Subgraph}

\wx{
Formally, we can construct the explanatory subgraph $\Set{G}^{*}_{K}$ by maximizing an attribution metric $A(\cdot)$:
\begin{gather}\label{equ:optimization-causal-attribution-of-subgraph}
    \Set{G}^{*}_{K}=\arg\max_{\Set{G}_{K}\subseteq\Set{G}}A(\Set{G}_{K}|\hat{y}_{c}),
\end{gather}
which is optimized over all possible combinations of $K$ edges: $\{\Set{G}_{K}|\Set{G}_{K}\subseteq\Set{G},|\Set{G}_{K}|=K\}$;
$A: \Space{G}\rightarrow\Space{R}$ measures the contribution of each candidate subgraph $\Set{G}_{K}$ to the target prediction $\hat{y}_{c}=f(\Set{G})$.

Towards causal explainability, $A(\Set{G}_{K}|\hat{y}_{c})$ needs to quantify the causal effect of $\Set{G}_{K}$.
We can directly manipulate the value of the input graph variable and investigate how the model prediction would be going on.
Such an operation is the intervention in causal inference \cite{pearl2016causal,pearl2009causality}, which is founded on the $do(\cdot)$ calculus.
It cuts off all the incoming links of a variable and forcefully assigns a variable with a certain value, which is no longer affected by its causal parents.}
For instance, $do(G=\Set{G})$ fixes the value of graph variable $G$ as a specific graph instance $\Set{G}$, short for $do(\Set{G})$.

\wx{
Through interventions, we formulate individual causal effect (ICE) \cite{pearl2016causal,pearl2009causality} in the attribution function $A(\Set{G}_{K}|\hat{y}_{c})$.
In particular, we view the input graph variable $G$ as our control variable and conduct two interventions: $do(G=\Set{G}_{K})$ and $do(G=\emptyset)$, which separately indicate} that the input receives treatment (\ie feeding $\Set{G}_{K}$ into the GNN) and control (\ie feeding uninformative reference into the GNN).
The ICE is the difference between potential outcomes under treatment and control: $Y(do(\Set{G}_{K}))-Y(do(\emptyset))$, where $Y$ is the prediction variable.
However, this difference does leaves the target prediction $\hat{y}_{c}$ untouched, thus easily distills degenerated explanation from $\Set{G}$.
Hence, we introduce a variable $I(G;Y)$ to estimate the mutual information \cite{DBLP:books/wi/01/CT2001,DBLP:conf/icml/BelghaziBROBHC18} between $G$ and $Y$, \wx{which is defined as:
\begin{gather}
    I(G;Y) = H(Y) - H(Y|G),\nonumber
\end{gather}
where $H(Y)$ and $H(Y|G)$ are the entropy and conditional entropy terms, respectively.
Involving the target prediction $\hat{y}_{c}$, we reformulate the ICE of $\Set{G}_{K}$ as:
\begin{gather}\label{equ:causal-attribution-of-subgraph}
    A(\Set{G}_{K}|\hat{y}_{c}) = I(do(\Set{G}_{K});\hat{y}_{c})-I(do(\emptyset);\hat{y}_{c}),
\end{gather}
where $I(do(\Set{G}_{K});\hat{y}_{c})=I(\Set{G}_{K};\hat{y}_{c})$} is to quantify the amount of information pertinent to the target prediction $\hat{y}_{c}$ held in the intervened graph $\Set{G}_{K}$;
analogously, $I(do(\emptyset);\hat{y}_{c})=I(\emptyset;\hat{y}_{c})$.

\vspace{5pt}
\noindent\textbf{Limitations.}
Nonetheless, directly optimizing Equation~\eqref{equ:optimization-causal-attribution-of-subgraph} faces two obstacles:
(1) This optimization is generally NP-hard \cite{L2X,DBLP:conf/iclr/ZhuNC20}, since the constraint $|\Set{G}_{K}|=K$ casts the task as a combinatorial optimization problem, where the number of possible subgraphs $\{\Set{G}_{K}\subseteq\Set{G}\}$ is super-exponential with the number of edges;
and (2) it only exhibits the contribution of a subgraph to the prediction holistically.
However, highlighting the importance of each componential edge is more preferred than showing a holistic subgraph solely.

\subsection{Causal Screening of An Edge Sequence}

To address these limitations, we propose the causal screening strategy to assess the causal effect of an edge sequence instead.
The basic idea is integrating the screening strategy \cite{aliferis2010local} with the cause-effect estimation of edges.
Specifically, the explanatory subgraph starts from the empty set, and incorporates the salient edges incrementally, one edge at a time.
Formally, the objective function is:
\begin{gather}\label{equ:edge-optimization}
    e^{*}_{k} = \arg\max_{e_{k}\in\Set{O}_{k}}A(e_{k}|\Set{G}^{*}_{k-1},\hat{y}_{c}),\quad k=1,2,\cdots,K,
\end{gather}
where $\Set{G}^{*}_{k-1}=\{e^{*}_{1},\cdots,e^{*}_{k-1}\}$ is the set of the first $(k-1)$ added edges after $(k-1)$ steps, and $\Set{G}_{0}=\emptyset$ at the initial step;
at the $k$-th step, $e^{*}_{k}$ is the edge selected from the pool of edge candidates $\Set{O}_{k}=\Set{G}\setminus\Set{G}^{*}_{k-1}$.
\wx{Having established $\Set{G}^{*}_{k}$ by merging $\Set{G}^{*}_{k-1}$ and $e^{*}_{k}$, we repeat this procedure until finding all top-$K$ edges as the final explanatory subgraph.}

\wx{
At each step, $A(e_{k}|\Set{G}^{*}_{k-1},\hat{y}_{c})$ guides the edge selection, which estimates the causal effect of $e_{k}$, conditioning on the previously selected edges $\Set{G}^{*}_{k-1}$.}
Specifically, given $\Set{G}^{*}_{k-1}$, we perform two interventions:
$do(\Set{G}^{*}_{k-1}\cup\{e_{k}\})$ represents that the input graph receives treatment (\ie the GNN takes the combination of $e_{k}$ and $\Set{G}^{*}_{k-1}$ as input);
whereas, $do(\Set{G}^{*}_{k-1})$ denotes that the input graph is under control (\ie the GNN takes $\Set{G}^{*}_{k-1}$ solely).
Afterward, we define the ICE of $e_{k}$ as:
\wx{\begin{align}\label{equ:causal-attribution-of-edge}
    A(e_{k}&|\Set{G}^{*}_{k-1},\hat{y}_{c})\nonumber\\
    &= I(do(\Set{G}^{*}_{k-1}\cup\{e_{k}\});\hat{y}_{c}) - I(do(\Set{G}^{*}_{k-1});\hat{y}_{c})\nonumber\\
    &= - H(\hat{y}_{c}|\Set{G}^{*}_{k-1}\cup\{e_{k}\}) + H(\hat{y}_{c}|\Set{G}^{*}_{k-1})\nonumber\\
    &= -p_{\theta}(\hat{y}_{c}|\Set{G})\log{\frac{p_{\theta}(\hat{y}_{c}|\Set{G}^{*}_{k-1})}{p_{\theta}(\hat{y}_{c}|\Set{G}^{*}_{k-1}\cup\{e_{k}\})}},
\end{align}
where $H(\hat{y}_{c}|\Set{G}^{*}_{k-1})=-p_{\theta}(\hat{y}_{c}|\Set{G})\log{p_{\theta}(\hat{y}_{c}|\Set{G}^{*}_{k-1})}$ is the term of conditional entropy; analogously, $H(\hat{y}_{c}|\Set{G}^{*}_{k-1}\cup\{e_{k}\})=-p_{\theta}(\hat{y}_{c}|\Set{G})\log{p_{\theta}(\hat{y}_{c}|\Set{G}^{*}_{k-1}\cup\{e_{k}\})}$; 
$p_{\theta}(\hat{y}_{c}|\Set{G})$, $p_{\theta}(\hat{y}_{c}|\Set{G}^{*}_{k-1})$, and $p_{\theta}(\hat{y}_{c}|\Set{G}^{*}_{k-1}\cup\{e_{k}\})$ are the prediction probabilities of the target class $\hat{y}_{c}$, when feeding $\Set{G}$, $\Set{G}^{*}_{k-1}$, and $\Set{G}^{*}_{k-1}\cup\{e_{k}\}$ into the model $f$, respectively.}
As a result, Equation \eqref{equ:causal-attribution-of-edge} explicitly assesses $e_{k}$'s interactions with the previously added edges.
One positive influence will be assigned with $e_{k}$, if it can collaborate with $\Set{G}^{*}_{k-1}$ and pursue a prediction more similar with $\hat{y}_{c}$; otherwise, a negative influence indicates that $e_{k}$ is not suitable to participate in the interpretation at step $k$.
\wx{Hence, it allows us to answer causality-related questions like “Given the previously selected edges, what is the causal effect of an edge on the target prediction?”.}

One straightforward solution to optimizing Equation~\eqref{equ:edge-optimization} is through greedy sequential exhaustive search.
One step consists of first calculating the ICE scores of all edge candidates, and then adding the edge with the most significant score to connect the previously added edges.
This step is iteratively repeated for $K$ times.
The greedy sequential exhaustive search is at the core of many feature selection algorithms~\cite{IFBS}, having shown great success.

\vspace{5pt}
\noindent\textbf{Limitations.}
However, there are two inherent limitations in the exhaustive search:
(1) This parameter-free approach explains every graph instance individually, hence is insufficient to offer a global understanding of the GNN, such as class-wise explanations~\cite{PGExplainer,ReFine}.
(2) The computational complexity of interpreting a graph is $O(2(|\Set{G}|-K)\times K/2))$.
This will be a bottleneck when explaining large-scale graphs that involve massive edges like social networks.
As a result, these limitations hinder the exhaustive search from being efficiently and widely used in real-world scenarios.

\wx{
\subsection{Reinforced Causal Explainer (RC-Explainer)}
To remedy the limitations, we revisit the sequential selection of edges from the viewpoint of reinforcement learning (RL) \cite{GCPN,AlphaGo}.
We then devise a RL agent, RC-Explainer, whose policy network learns how to conduct the causal screening.
As the policy network explores the post-hoc explanations over the population of all training graphs, RC-Explainer is able to systematize the global view of important patterns and hold the global understanding of the model's workings.
Moreover, benefiting from the RL scheme, RC-Explainer can efficiently determine the edge sequence with the significant causal attributions.

\subsubsection{\textbf{Causal Screening as Reinforcement Learning}}
\label{sec:rl-environment}
Following previous work \cite{GCPN}, we frame the causal screening process as a Markov Decision Process (MDP) $M=\{\Set{S},\Set{A},P,R\}$.
Herein, $\Set{S}=\{s_{k}\}$ is the set of states abstracting edge sequences during exploration,
and $\Set{A}=\{a_{k}\}$ is the set of actions, which adds an edge to the current edge sequences at each step.
Under the Markov property \cite{GCPN,CGVAE}, $P(s_{k}|s_{k-1},a_{k})$ is the transition dynamics that specifies the new edge sequence $s_{k}$ after making an action of edge addition $a_{k}$ on the prior state $s_{k-1}$.
$R(s_{k-1},a_{k})$ is to quantify the reward after making the action $a_{k}$ from the prior state $s_{k-1}$.
As such, the trajectory $(s_{0},a_{1},r_{1},s_{1},\cdots,a_{K},r_{K},s_{K})$ naturally depicts the generation of an edge sequence, where the step-wise reward $r_{K}$ reflects $a_{K}$'s causal effect and coalition effect with the previously-added edges.
Note that, various graphs with the model predictions describe different environments being explored by the RL agent.
Here we elaborate the foregoing key elements for one target graph $\Set{G}$ and its prediction $\hat{y}_{c}$ as follows.

\vspace{5pt}
\noindent\textbf{State Space.}
At step $k$, the state $s_{k}$ indicates the subgraph $\Set{G}_{k}$ consisting of the explored edges: $s_{k}=\Set{G}_{k}$, where the initial state $s_{0}=\Set{G}_{0}=\emptyset$.
Figure \ref{fig:framework} illustrates the changes in varying states.

\vspace{5pt}
\noindent\textbf{Action Space.}
Observing the state $s_{k-1}=\Set{G}_{k-1}$, the available action space $\Set{A}_{k}$ is the complement of $\Set{G}_{k-1}$, formally $\Set{A}_{k}=\Set{G}\setminus\Set{G}_{k-1}$.
The RL agent picks up a salient edge $e_{k}$ from $\Set{A}_{k}$ and connects it to the previous selection $\Set{G}_{k-1}$. That is, this action of edge addition can be represented as $a_{k}=e_{k}$ for simplicity.

\vspace{5pt}
\noindent\textbf{State Transition Dynamics.}
Having made the action $a_{k}=e_{k}$ at step $k$, the transition of the state $s_{k}$ is merging $e_{k}$ into the previous state $s_{k-1}$: $\Set{G}_{k}=\Set{G}_{k-1}\cup\{e_{k}\}$.

\vspace{5pt}
\noindent\textbf{Reward Design.}
To measure the quality of the action $a_{k}$ at step $k$ and guide the further explorations, we consider two factors into the reward design:
\begin{itemize}
    \item Validity of action $a_{k}=e_{k}$, which reflects the ICE of $e_{k}$ (\cf Equation \eqref{equ:causal-attribution-of-edge}). A positive attribution suggests that the newly-added edge $e_{k}$ contributes uniquely and positively to the explanation --- that is, it not only serves as the plausible causal determinant to the target prediction $\hat{y}_{c}$, but also cooperates with the previous edges $\Set{G}_{k-1}$ as an effective coalition.
    In contrast, a vanishing (near-to-zero) or negative attribution reveals that the edge is redundant to the previous edges or fails to explain the target prediction.
    
    \item Validity of state $s_{k}=\Set{G}_{k}$, which justifies the predictive ability of the current explanatory subgraph.
    Here we assign a positive score ($1$) to $\Set{G}_{k}$ that can successfully explain the target prediction, while we use a negative score ($-1$) to penalize $\Set{G}_{k}$ for the wrong prediction.
\end{itemize}
In a nutshell, the reward $R(s_{k-1},a_{k})$ for the intervention action $a_{k}=e_{k}$ at state $s_{k-1}=\Set{G}_{k-1}$ is formulated as:
\begin{gather}
    R(\Set{G}_{k-1},e_{k}) =
    \begin{cases}
        A(e_{k}|\Set{G}_{k-1},\hat{y}_{c})+1,\text{if}~f_{\theta}(\Set{G}_{k-1}\cup\{e_{k}\})=\hat{y}_{c}\\
        A(e_{k}|\Set{G}_{k-1},\hat{y}_{c})-1,\text{otherwise.}
    \end{cases}\nonumber
\end{gather}
}

\wx{
\subsubsection{\textbf{Policy Network}}
Having outlined the causal screening environment, we now present the policy network $q_{\phi}$ of RC-Explainer to explore in the environment.
Specifically, it takes the pair of intermediate state $G_{k-1}$ and the target prediction $\hat{y}_{c}$ as the input, and aims to determine the next action $e_{k}$:
\begin{gather}
    e_{k}\sim q_{\phi}(\Set{G}_{k-1},\hat{y}_{c}),
\end{gather}
where $\phi$ summarizes the trainable parameters of the policy network; $e_{k}$ is yielded with the probability $P_{\phi}(e_{k}|\Set{G}_{k-1},\hat{y}_{c})$ of being added to the explanatory subgraph.

\vspace{5pt}
\noindent\textbf{Representation Learning of Action Candidates.}
With the space of action candidates $\Set{A}_{k}=\Set{G}\setminus\Set{G}_{k-1}$, our policy network first learns the representation for each action candidate $a_{k}\in\Set{A}_{k}$.
As introduced in Section \ref{sec:rl-environment}, $a_{k}$ is performed on the edge $e_{k}$, whose two endpoints are denoted by $u$ and $v$, \ie $e_{k}=(v_,u)$.
We employ another GNN model $g$ over the full graph $\Set{G}$ to create the node representations of $u$ and $v$, and then combine them together to obtain the action representation of $e_{k}$ as follows:
\begin{gather}
    \Mat{z}_{e_{k}} = \text{MLP}_{1}([\Mat{z}_{v}||\Mat{z}_{u}||\Mat{x}_{e_{k}}]),
\end{gather}
where $g$ yields $\Mat{z}_{v}\in\Space{R}^{d'}$ and $\Mat{z}_{u}\in\Space{R}^{d'}$ to separately represent $v$ and $u$ (\cf Equation \eqref{equ:gnn-graph-representation});
$\Mat{x}_{e}\in\Space{R}^{d_{2}}$ is the pre-existing feature of $e_{k}$, which can be ignored when no feature is available;
$\cdot||\cdot$ is the concatenation operator.
We use a MLP with one hidden layer $\Mat{W}^{(2)}\sigma(\Mat{W}^{(1)}[\Mat{z}_{v}||\Mat{z}_{u}||\Mat{x}_{e_{k}}])$ to obtain $\Mat{z}_{e_{k}}\in\Space{R}^{d''}$, where $\sigma$ is a ReLU nonlinearity.
Note that the model parameters $\mu$ of $g$ are trainable, while the model parameters $\theta$ of the target GNN model $f$ are fixed.

\vspace{5pt}
\noindent\textbf{Selection of Action.}
Having established the representations of action candidates, we aim to select one action from the space and perform it.
Instead of trying candidates exhaustively, the policy network learns the importance of making an action $a_{k}=e_{k}$ to the current state $s_{k-1}=\Set{G}_{k-1}$:
\begin{gather}
    p_{e_{k}}=\text{MLP}_{2,c}([\Mat{z}_{e_{k}}||\Mat{z}_{\Set{G}_{k-1}}]),
\end{gather}
where $\Mat{z}_{\Set{G}_{k-1}}$ is the representation of the current explanatory subgraph $\Set{G}_{k-1}$, which aggregates the representations of its compositional nodes via Equation \eqref{equ:gnn-graph-representation}.
We use a class-specific MLP with one hidden layer $\Mat{W}^{(4,c)}\sigma(\Mat{W}^{(3)}[\Mat{z}_{e_{k}}||\Mat{z}_{\Set{G}_{k-1}}])$ to get the scalar $p_{e_{k}}$, where $c$ corresponds to the target class $\hat{y}_{c}$.

Thereafter, we apply a softmax function over all action candidates $\Set{A}_{k}$ to convert their importance scores into the probability distribution.
Formally, the probability of $e_{k}$ being selected as the action is as:
\begin{gather}
    P_{\phi}(e_{k}|\Set{G}_{k-1},\hat{y}_{c})=\text{SOFTMAX}_{\Set{A}_{k}}(p_{e_{k}}),
\end{gather}
where $\phi$ collects parameters of $g_{\mu}$, MLP$_{1}$, and $\{\text{MLP}_{2,c}\}_{c=1}^{C}$.
It is worth mentioning that the class-wise MLPs latch on class-wise knowledge across the training graphs with the same prediction.

\subsubsection{\textbf{Policy Gradient Training}}
However, SGD cannot be directly used for the optimization, since the discrete sampling within the policy network blocks gradients.
To solve this issue, we adopt the policy gradient-based training framework, REINFORCE \cite{REINFORCE,GCPN}, to optimize the policy network as follows:
\begin{gather}
    \max_{\phi}\Space{E}_{\Set{G}\in\Set{O}}\Space{E}_{k}[R(\Set{G}_{k-1},e_{k})\log{P_{\phi}(e_{k}|\Set{G}_{k-1},\hat{y}_{c})}].
\end{gather}
Obviously, this training framework encourages the actions with large rewards --- sequentially finding the edges with the salient causal attributions to construct the explanatory subgraph.

As a result, our RC-Explainer inherits the desired characteristics of causal screening, which considers the causal effects of edges and the dependencies among edges.
Furthermore, RC-Explainer is optimized on all graphs in the training set; thus, it holds a global view of the target model's inner workings and achieves better generalization ability to generate explanations for unseen graph instances.

}

\subsection{Discussion}
\subsubsection{\textbf{Time Complexity}}
When explaining one graph $\Set{G}$, the time cost mainly comes from the two components: (1) the representation learning of node representations, (2) the step-wise representation learning of edge candidates and action prediction.
Specifically, using a trainable GNN model $g_{\mu}$ to generate node representations has computational complexity $O(\sum_{l=1}^{L}|\Set{G}|\times d_{l}\times d_{l-1})$, where $d_{l}$ is the representation dimension at $l$-th layer.
At each step $k$, the time cost is $O(|\Set{A}_{k}|\times 2d'\times d'')$ for creating representations of edge candidates, while being $O(|\Set{A}_{k}|\times d'')$ for predicting one action.
As a result, the cost of generating the explanation $\Set{G}_{K}$ is $O(\sum_{k=1}^{K}|\Set{A}_{k}|\times (2d'\times d''+d''))$.
In total, the time complexity of the whole training episode is $O(\sum_{\Set{G}\in\Set{O}}(\sum_{l=1}^{L}|\Set{G}|\times d_{l}\times d_{l-1}+\sum_{k=1}^{K}|\Set{A}_{k}|\times (2d'\times d''+d'')))$, where $\Set{O}$ is the training set.

\wx{
\subsubsection{\textbf{Potential Limitations}}
We list two potential limitations RC-Explainer might face, expensive time complexity and out-of-distribution issue:
\begin{itemize}
    \item Although RC-Explainer benefits from reinforcement learning and is more efficient than the exhaustive search, it might still suffer from the high computational cost, when generating the sampling probabilities over the large action space.
    See Section \ref{sec:evaluation-time} for the evaluations \wrt empirical time complexity.
    It limits RC-Explainer's development on large scale graphs.
    Hence we leave the pruning of action space and the improvement of scalability in future work, such as pre-judging the probability of edges being selected and extracting a small set of edges as the action candidates.

    \item The causal attributions of holistic subgraph (\cf Equation \eqref{equ:causal-attribution-of-subgraph}) and edge sequence (\cf Equation \eqref{equ:causal-attribution-of-edge}) are founded upon on the feature removal principle \cite{FeatureRemoval} --- that is, the complement of the subgraph is discarded, while the target model takes the subgraph as the input only.
    Nonetheless, feature removal makes subgraphs off the distribution of the full graphs, thus causing the out-of-distribution (OOD) issue.
    The OOD subgraph hardly follows the degree distribution \cite{DBLP:conf/kdd/LeskovecKF05}, graph sizes \cite{DBLP:conf/icml/BevilacquaZ021} or possibly violates constraints \cite{CGVAE}.
    As a result, the OOD issue could bring a spurious correlation between the true importance of subgraph and the model prediction, making the ICE estimation unfaithful and unreliable.
    See Section \ref{sec:qualitative-analyses} for the failure cases, where the explanations suffer from the OOD effect.
    We will explore the solution to the OOD issue in future work, such as using the counterfactual generation to fulfil the subgraph and make it conform to the original distribution
\end{itemize}


}
\section{Experiments}
To demonstrate the effectiveness of RC-Explainer, we investigate the following research question: Can RC-Explainer provide more reasonable explanations than the state-of-the-art explainer methods?

\begin{table}[t]
    \caption{Dataset statistics with model configurations.}
    \vspace{-5pt}
    \label{tab:data-gnn-statistics}
    \resizebox{0.96\columnwidth}{!}{
    \begin{tabular}{r|c|c|c}
    \hline
     & Mutagenicity & REDDIT-MULTI-5K & Visual-Genome\\ \hline\hline
    Graphs\# & 4,337 & 4,999 & 4,443 \\
    Classes\# & 2 & 5 & 5  \\
    Avg. Nodes\# & 30.32 & 508.52 & 35.32 \\
    Avg. Edges\# & 30.77 & 594.87 & 18.04 \\ \hline\hline
    Target GNNs & GIN & k-GNN & APPNP \\ 
    Layers\# & 2 & 3 & 2 \\ 
    Accuracy & 0.806 & 0.644 & 0.640 \\ \hline
    \end{tabular}}
    \vspace{-10pt}
\end{table}

\begin{figure*}[t]
	\centering
	\subcaptionbox{Mutagenicity\label{fig:gamma-hit-mi}}{
		\includegraphics[width=0.3\textwidth]{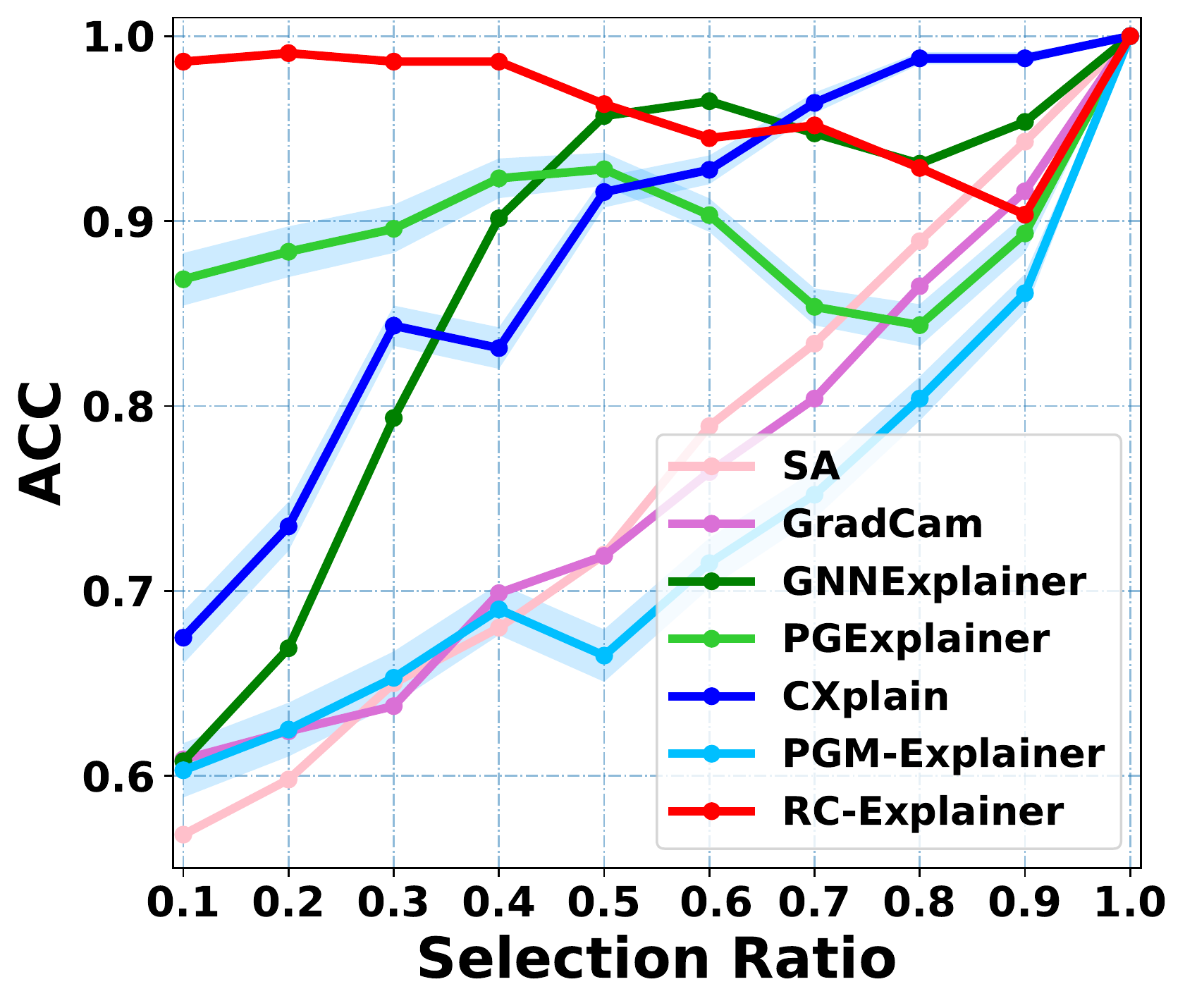}}
	\subcaptionbox{REDDIT-MULTI-5K\label{fig:gamma-ndcg-mi}}{
        \includegraphics[width=0.3\textwidth]{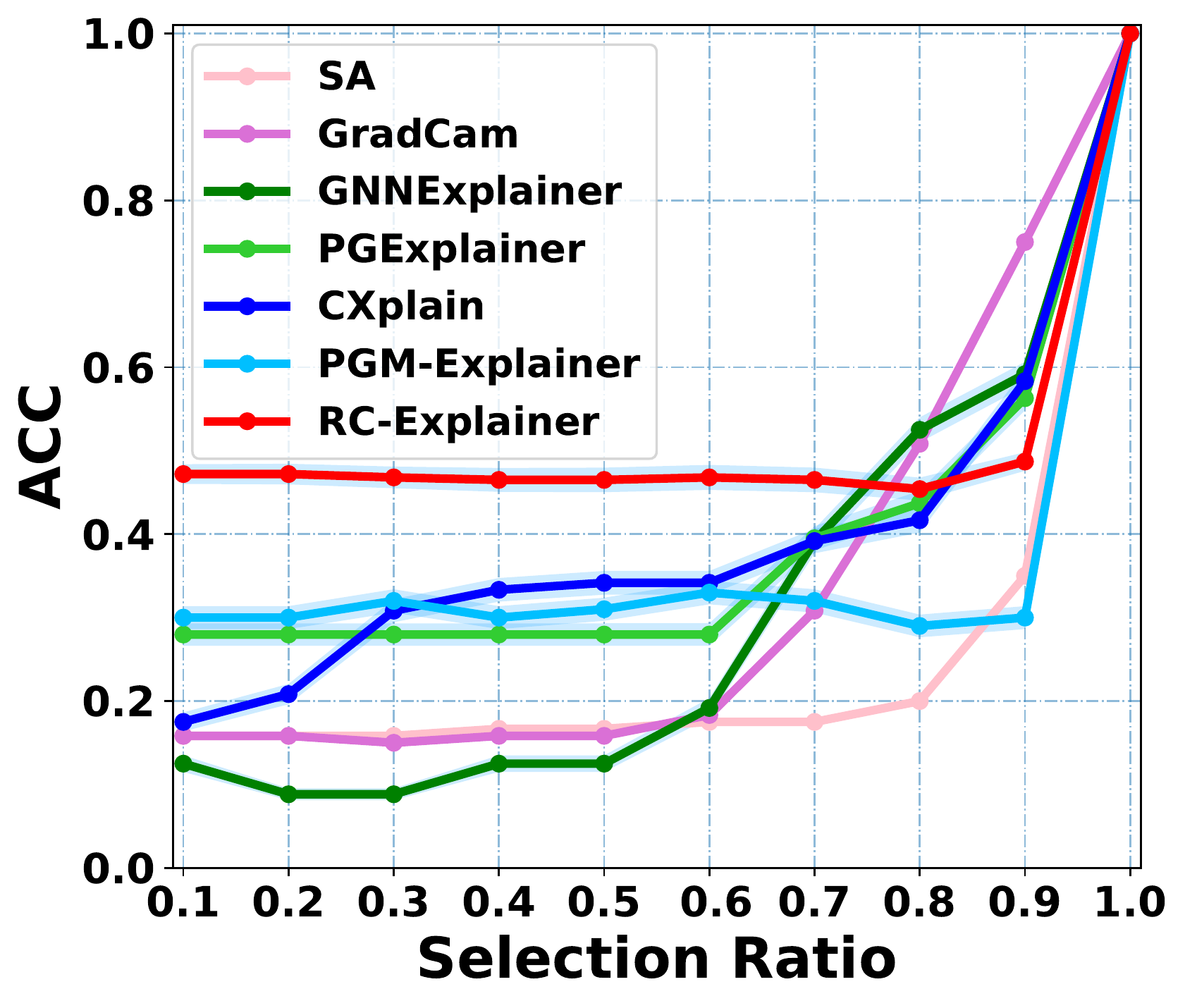}}
    \subcaptionbox{Visual Genome\label{fig:gamma-ndcg-mi}}{
        \includegraphics[width=0.3\textwidth]{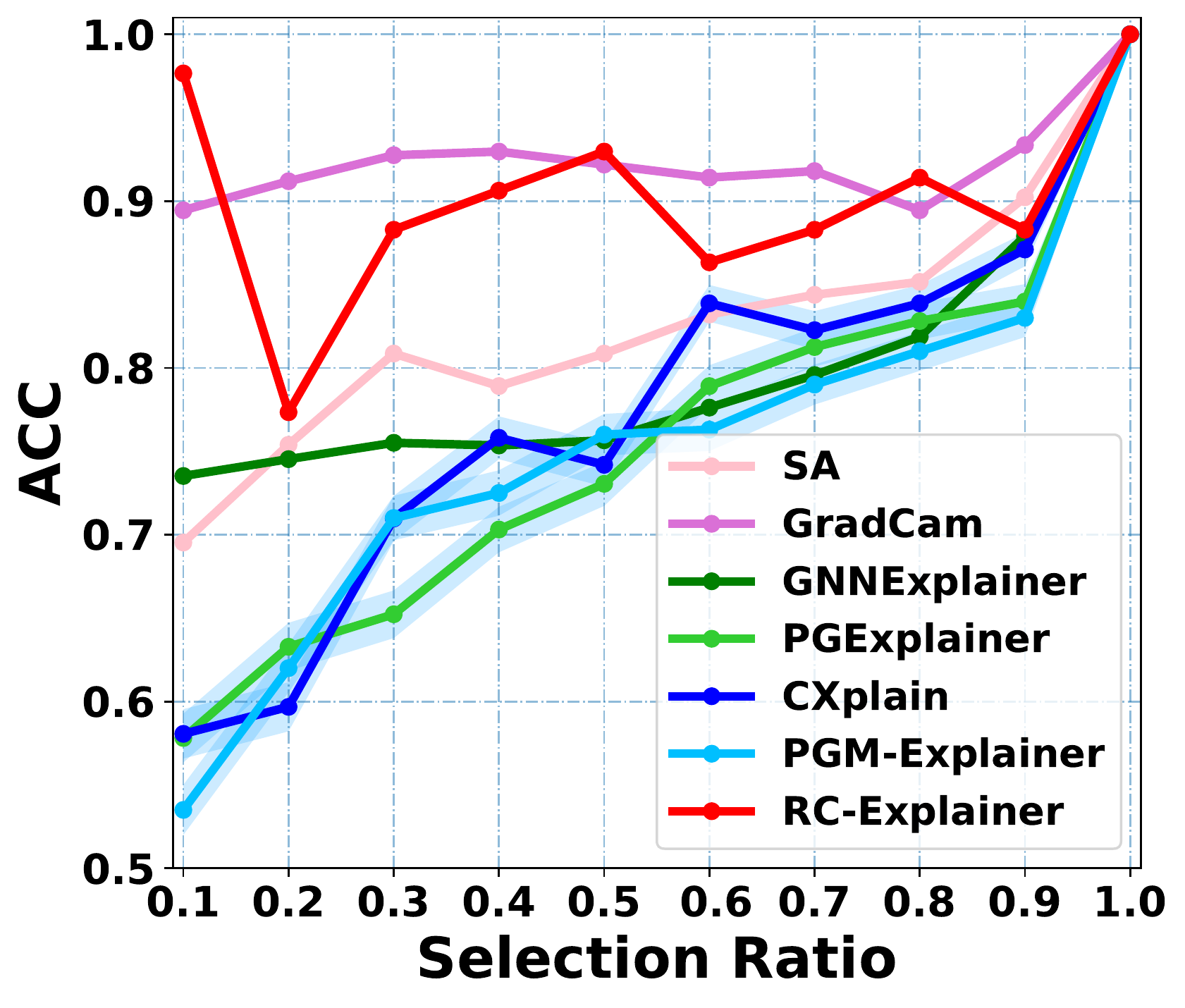}}
	\vspace{-5pt}
	\caption{Accuracy curves of all explainers over different selection ratios. Best viewed in color.}
	\label{fig:acc-curve}
	\vspace{-15pt}
\end{figure*}

\subsection{Experimental Settings}
\subsubsection{\textbf{Dataset Description.}}
We use three benchmark datasets in the experiments:
\begin{itemize}
    \item \textbf{Mutagenicity}~\cite{kazius2005derivation,riesen2008iam} has $4,377$ molecule graphs, where each graph is labeled with one of two labels: mutagenic and non-mutagenic, based on the mutagenic effect on a bacterium. We trained a GIN model \cite{DBLP:conf/iclr/XuHLJ19,DBLP:conf/iclr/HuLGZLPL20} to perform the binary classification.
    \item \textbf{REDDIT-MULTI-5K}~\cite{DBLP:conf/kdd/YanardagV15} has $4,999$ social networks labeled with five different classes to indicate the topics of question/answer communities. Upon them, we trained a k-GNN model \cite{DBLP:conf/aaai/0001RFHLRG19} as a classifier.
    \item \textbf{Visual Genome}~\cite{DBLP:journals/ijcv/KrishnaZGJHKCKL17} is a collection of images. Each image is coupled with a scene graph, where nodes are the objects with bounding boxes and edges are the relations between objects. Following the previous work~\cite{Grad-CAM-Graph}, we extract $4,443$ (images, scene graphs) pairs covering five classes: stadium, street, farm, surfing, forest. A GNN model, APPNP~\cite{APPNP}, is trained as a classifier, where the regional images are treated as the node features.
\end{itemize}
Table \ref{tab:data-gnn-statistics} shows the statistics of datasets with the configurations of GNN models.
We follow prior studies \cite{Grad-CAM-Graph,GNNExplainer} and partition each dataset into training, validation, and testing sets with the ratio of $80\%$:$10\%$:$10\%$.
Having obtained the trained GNNs on the training sets, we train the parametric explainers on the training sets, and tune their hyper-parameters on the validation sets.
We run the explainers five times on the testing sets to report the average results.

\subsubsection{\textbf{Evaluation Metrics.}}
It is of crucial importance to evaluate the explanations quantitatively.
However, the ground-truth knowledge of explanations is usually unavailable in real-world datasets.
Hence, many prior works \cite{L2X,DBLP:conf/kdd/LiangBCBW20,SanityCheck,Towardsbetter} have proposed some metrics to assess the explanations without the ground truth.
Here we adopt two widely-adopted metrics:
\begin{itemize}
    \item \textbf{Predictive Accuracy}~\cite{L2X,DBLP:conf/kdd/LiangBCBW20} (ACC@$\mu$) is to measure whether using the explanatory subgraph can successfully recover the target prediction:
    \wx{
    \begin{gather}
        \text{ACC($\mu$)}=\Space{E}_{\Set{G}\sim\Space{G}}[\Space{I}(f(\Set{G}),f(\Set{G}^{*}_{K}))],
    \end{gather}
    where $\mu$ is the selection ratio (\eg $5\%$), $K= \lceil \mu\times|\Set{G}|\rceil$ is the size of explanatory subgraph; $\Space{I}(\cdot)$ is the indicator function to check whether $f(\Set{G})$ equals to $f(\Set{G}^{*}_{K})$. Moreover, we also report the ACC curve over different selection ratios $[0.1, 0.2, \cdots, 1.0]$ and denote the area under curve as ACC-AUC.}
    
    \item \textbf{Contrastivity}~\cite{Grad-CAM-Graph} (CST) quantifies the invariance between class-discriminative explanations.
    \wx{It is built upon the intuition that} a reasonable method should yield differing explanations, in response to the target prediction changes. \wx{Here we formulate it as} the Spearman rank correlation between class-discriminative edge scores:
    \wx{
    \begin{gather}
        \text{CST}=\Space{E}_{\Set{G}\sim\Space{G}}\Space{E}_{s\neq\hat{y}}[|\rho(\Phi(\Set{G},s), \Phi(\Set{G},\hat{y}))|],
    \end{gather}
    where $s$ means we permute the label $\hat{y}$ of $\Set{G}$ being interpreted; $\Phi(\Set{G},\hat{y})$ is the attribution scores of all edges; $|\rho(\cdot)|$ is the Spearman rank correlation with the absolute value to measure the invariance between the edge attributions, in response to the label changes.
    }
\end{itemize}
Moreover, as suggested in the prior study \cite{Towardsbetter}, the explanations generated by a reasonable explainer should be dependent on the target model --- that is, presenting a faithful understanding of how the target model works.
Hence we also conduct sanity check:
\begin{itemize}
    \item \textbf{Sanity Check} on model randomization is to compare the attribution scores on the trained GNN $f$ with that on an untrained GNN $\hat{f}$ with randomly-initialized parameters.
    Similar attributions infer that the explainer is insensitive to the model changes, thus fails to pass the check.
    \wx{Here we frame it as the rank correlation between these two attributions:
    \begin{gather}
        \text{SC}=\Space{E}_{\Set{G}\sim\Space{G}}[|\rho(\Phi(\Set{G},f(\Set{G})), \Phi(\Set{G},\tilde{f}(\Set{G})))|].
    \end{gather}
    Similar attributions infer the explainer is insensitive to properties of the model, thus failing to pass the check.}
\end{itemize}

\subsubsection{\textbf{Baselines.}}
We compare RC-Explainer with the state-of-the-art explainers, covering the gradient-based (SA \cite{SA-Graph}, Grad-CAM\cite{Grad-CAM-Graph}), masking-based (GNNExplainer \cite{GNNExplainer}), attention-based (PGExplainer \cite{PGExplainer}), and perturbation-based (CXPlain \cite{CXPlain}, PGM-Explainer\cite{PGM-Explainer}).

\begin{table*}[t]
    \centering
    \caption{Predictive accuracy of explanations derived from explainers. The best performance is highlighted with $^{*}$, while the second-best performance is underlined.}
    \vspace{-5pt}
    \label{tab:overall-performance-acc}
    \resizebox{0.96\textwidth}{!}{
    \begin{tabular}{cc||cccccc||c}
    \hline
     &  & SA & Grad-CAM & GNNExplainer & PGExplainer & CXPlain & PGM-Explainer & RC-Explainer \\ \hline\hline
    
     \multirow{3}{*}{ACC@$10\%\uparrow$} & Mutagenicity & 0.568 & 0.608 &  0.607 & 0.674 & \underline{0.868} & 0.603 & \textbf{0.986}$^{*}$ \\ 
     & REDDIT-MULTI-5K & 0.158 & 0.158 & 0.125 & 0.175 & 0.279 & \underline{0.300} & \textbf{0.472}$^{*}$ \\ 
     & Visual Genome & 0.695 & \underline{0.894} & 0.735 & 0.580 & 0.578 & 0.535 & \textbf{0.976}$^{*}$ \\\hline\hline

     \multirow{3}{*}{ACC-AUC$\uparrow$} & Mutagenicity & 0.767 & 0.764 & 0.872 & \underline{0.899} & 0.886 & 0.737 & \textbf{0.964}$^{*}$ \\
     & REDDIT-MULTI-5K & 0.362 & 0.399 & 0.429 & 0.412 & \underline{0.439} & 0.377 & \textbf{0.521}$^{*}$ \\ 
     & Visual Genome & 0.829 & \textbf{0.917}$^{*}$ & 0.802 & 0.757 & 0.775 & 0.754 & \underline{0.901} \\ \hline

    \end{tabular}}
    \vspace{-5pt}
\end{table*}

\begin{table*}[t]
    \centering
    \caption{Other quantitative analyses for explainers \wrt contrastivity metrics, sanity check, and time complexity. Symbol $(\cdot)$ indicates the rank of RC-Explainer over all methods.}
    \vspace{-5pt}
    \label{tab:overall-performance}
    \resizebox{0.96\textwidth}{!}{
    \begin{tabular}{cc||cccccc||c}
    \hline
     &  & SA & Grad-CAM & GNNExplainer & PGExplainer & CXPlain & PGM-Explainer & RC-Explainer \\ \hline\hline
     
     \multirow{3}{*}{CST$\downarrow$} & Mutagenicity & 0.975 & 0.768 & 0.690 & 0.202 & 0.587 & 0.343 & 0.311$^{(2)}$ \\ 
     & REDDIT-MULTI-5K & 0.513 & 0.211 & 0.945 & 0.145 & 0.664 & 0.061 & \wx{0.481$^{(4)}$} \\ 
     & Visual Genome & 0.462 & 0.426 & 0.417 & 0.421 & 0.320  & 0.403 & \wx{0.306$^{(1)}$} \\ \hline\hline

     \multirow{3}{*}{SC$\downarrow$} & Mutagenicity & 0.221 & 0.254 & 0.124 & 0.278 & 0.327 & 0.597 & 0.248$^{(3)}$ \\
     & REDDIT-MULTI-5K & 0.183 & 0.537 & 0.040 & 0.123 & 0.696 & 0.829 & 0.465$^{(4)}$ \\ 
     & Visual Genome & 0.321 & 0.375 & 0.676 & 0.831 & 0.266 & 0.588 & 0.309$^{(2)}$ \\ \hline\hline
     
     \multirow{3}{*}{\begin{tabular}[c]{@{}c@{}}Time\\ (per graph)\end{tabular}} & Mutagenicity & 0.011 & 0.010 & 2.57 & 0.026 & 1.74 & 1.19 &  0.680     \\
     & REDDIT-MULTI-5K & 0.008 & 0.008 & 2.14 & 0.021 & 13.4 & 64.2 &  23.8\\ 
     & Visual Genome & 0.015 & 0.015 & 2.71 & 0.028 & 1.77 & 1.58 &  0.339\\ \hline
     
    \end{tabular}}
    \vspace{-10pt}
\end{table*}

\subsubsection{\textbf{Parameter Settings.}}
To facilitate reproducibility, we release codes and datasets at \url{https://github.com/xiangwang1223/reinforced_causal_explainer}, and summarize the hyperparameter settings in the supplementary material.
For nonparametric methods (SA and Grad-CAM), we use the codes released by the original papers \cite{SA-Graph,Grad-CAM-Graph}.
For the other parametric methods, we conduct a grid search to confirm the optimal settings for each method.
To be more specific, the optimizer is set as Adam, the learning rate is tuned in $\{10^{-3},10^{-2},10^{-1}\}$, and the weight decay is searched in $\{10^{-5},10^{-4},10^{-3}\}$.
Other model-specific hyperparameters are set as follows:
For GNNExplainer, the weight of mutual information is fixed as $0.5$;
For PGExplainer, the temperature for reparameterization is $0.1$;
For PGM-Explainer, the number of perturbations is tuned in $\{10,100,1000\}$ with different perturbation modes.

\begin{figure*}[t]
    \centering
	\includegraphics[width=0.94\textwidth]{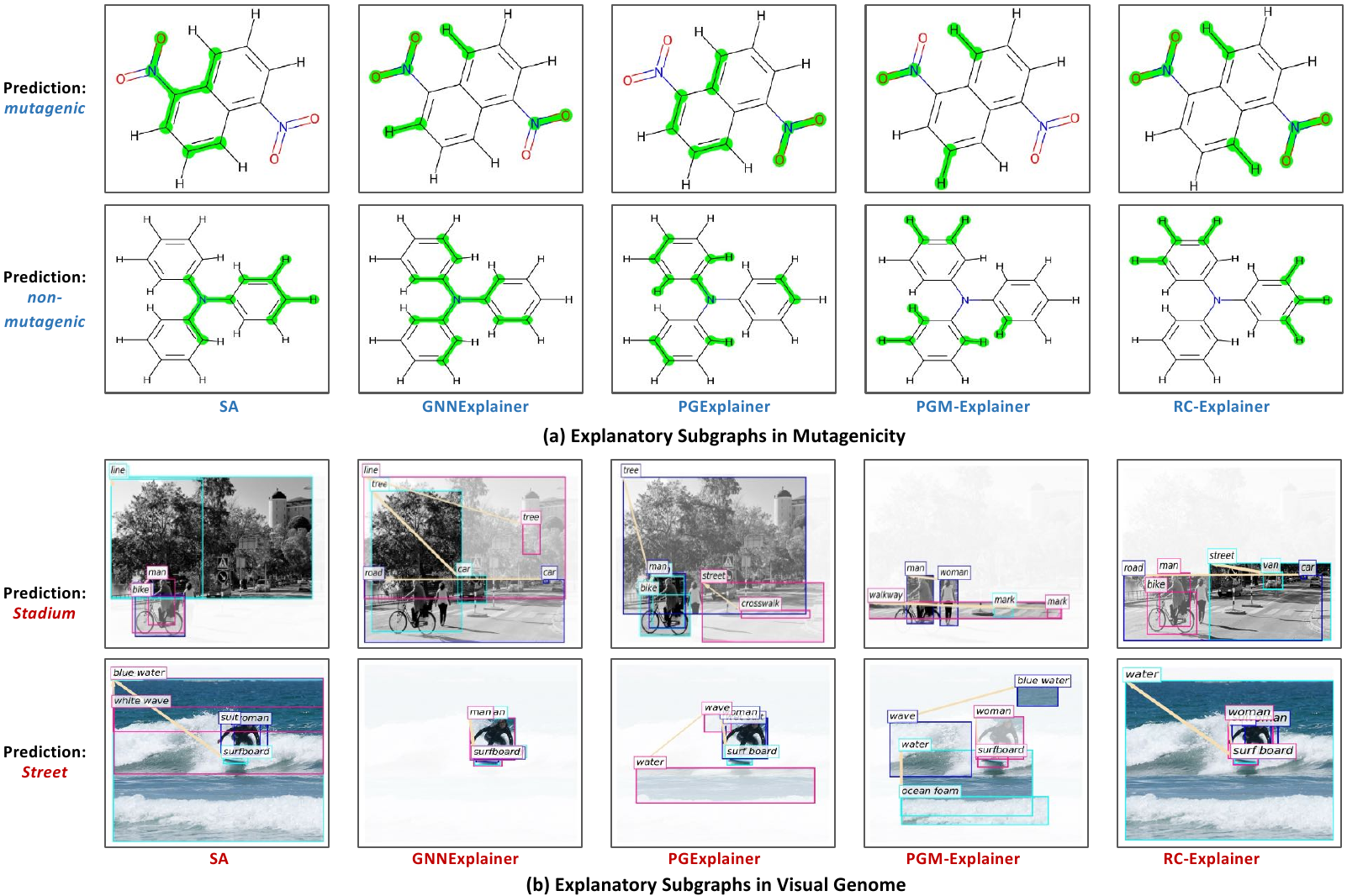}
	\vspace{-5pt}
	\caption{Selected explanations for each explainer, where the top $20\%$ edges are highlighted. Note that some edges have the same nodes. In Visual Genome, the objects involved in the edges are blurred based on the edge attributions; meanwhile, in Mutagenicity, a darker color of a bond indicates the larger attribution for the prediction.}
	\label{fig:case-study}
	\vspace{-10pt}
\end{figure*}

\subsection{Evaluation of Explanations}
\subsubsection{\textbf{Evaluation \wrt Predictive Accuracy}}
Table \ref{tab:overall-performance-acc} reports the empirical results \wrt ACC@$10\%$ and ACC-AUC, and Figure \ref{fig:acc-curve} demonstrates the ACC curves over different selection ratios.
We find that:
\begin{itemize}
    \wx{
    \item RC-Explainer outperforms the compared baselines by a large margin across all three datasets, when only $10\%$ of edges are selected as the explanatory subgraphs.
    For example, it achieves significant improvements over the strongest baselines \wrt ACC@$10\%$ by $13.59\%$ and $25.51\%$ in Mutagenicity and REDDIT-MULTI-5K, respectively.
    This verifies the rationality and effectiveness of RC-Explainer.
    We ascribe these improvements to two key characteristics of causal screening:
    (1) Assessing the causal effects of edges enables us to better distinguish the causal relationships from the spurious correlations between input edges and model outputs, thus encouraging causal explainability instead of statistical interpretability;
    (2) Benefiting from the screening strategy, RC-Explainer can reveal the dependencies of the candidate edges and the previous selections.
    It allows us to identify and explicit the coalition effect of edges, and reduce the redundancy in explanations.

    \item The explanatory subgraphs derived from RC-Explainer are influential to the model predictions. Especially, in Mutagenicity and REDDIT-MULTI-5K, RC-Explainer's ACC-AUC scores are $0.964$ and $0.901$, close to the optimal fidelity. This validates that RC-Explainer faithfully reflects the workings of the target GNNs.
    
    \item As Figure \ref{fig:acc-curve} shows, increasing the selection ratios might decrease RC-Explainer's predictive accuracy.
    This again justifies the rationality of our causal screening: when the causal determinants or coalitions have been selected, adding more noncausal or redundant edges has only a negligible impact on the accuracy.

    \item Analyzing the ACC curves and ACC-AUC scores in Visual Genome, we find that RC-Explainer only achieves comparable performance to Grad-CAM. 
    One possible reason is that node features could be more informative about scenes than the existence of edges. Taking Figure \ref{fig:case-study} as an example, the visual features of the \emph{road} node might be sufficient to predict the \emph{street} scene.
    Hence, Grad-CAM benefits from the context-enhanced gradients that capture the node features thus achieves high-quality explanations.
    It inspires us to further investigate the cooperation of node feature and graph structure in the explanation generation.
    
    \item In general, the line of causal explainability (\ie CXPlain, PGM-Explainer, RC-Explainer) performs better than the line of statistical explainability (\ie SA, Grad-CAM, GNNExplainer, PGExplainer). Because using gradients, masks, or attentions is prone to capturing the spurious input-output correlations and missing the causation.
    This is consistent with prior studies \cite{CXPlain,PGM-Explainer}.

    \item Within the line of causal explainability, RC-Explainer outperforms CXPlain and PGM-Explainer, suggesting that screening to combine an edge with the previous selection estimates the edges' causal effects more accurately. This emphasizes the effectiveness of considering edge dependencies.
    }
\end{itemize}

\subsubsection{\textbf{Evaluation \wrt Contrastivity}}
\wx{We move on to the contrastivity reported in Table \ref{tab:overall-performance} to check how explanations differ when the target predictions change.
We find that:
(1) Over all approaches, RC-Explainer achieves the lowest and the second-lowest CST scores in Visual Genome and Mutagenicity, respectively.
This indicates that the explanations of RC-Explainer are class-discriminative, thus offering an understanding of the decision boundary between different classes.
For instance, in REDDIT-MULTI-5K, different patterns of user behaviors can describe various topics of communities.
See Section \ref{sec:qualitative-analyses} for more supporting pieces of evidence;
(2) Jointly inspecting the ACC curves in Figure \ref{fig:acc-curve}, we observe that RC-Explainer will randomly pick up edges, once the most influential edges have been selected.
These randomly-added edges could increase the ranking correlations of class-specific explanations, thus influencing the overall CST scores of RC-Explainer.
}

\begin{figure*}[t]
    \centering
	\includegraphics[width=0.94\textwidth]{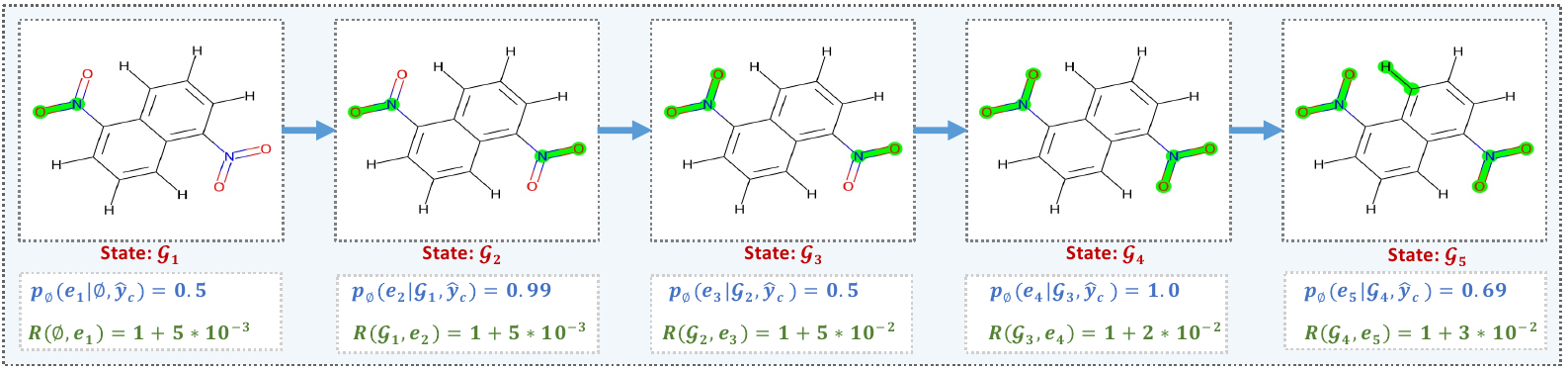}
	\vspace{-5pt}
	\caption{Showcasing the sequential decision process of explaining one graph in Mutagenicity.}
	\label{fig:sequence-show}
	\vspace{-10pt}
\end{figure*}

\subsubsection{\textbf{Evaluation \wrt Sanity Check}}
\wx{We then focus on the sanity checks presented in Table \ref{tab:overall-performance} and find that:
(1) RC-Explainer achieves relatively low SC scores and safely passes the checks, indicating that the attributions on the trained and untrained GNNs differ significantly.
This evidently shows that our feature attribution is dependent on the target GNN, which is necessary for faithfully interpreting the inner workings of the target model;
(2) Jointly analyzing the SC scores across three datasets, we find that RC-Explainer is less sensitive to the target model's status in Visual Genome.
We conjecture that the untrained GNN still works as the kernel function to filter some useful information, thus somehow reflects the edge importance.
We leave the justification of possible reasons for future work.
}

\subsubsection{\textbf{Evaluation \wrt Time Complexity}}
\label{sec:evaluation-time}

\wx{We present the actual runtime of all approaches in Table \ref{tab:overall-performance}, which reports the inference time to generate an explanation.
Our findings are:}
(1) In Mutagenicity and Visual Genome, RC-Explainer computes the attribution scores significantly faster than GNNExplainer and PGM-Explainer. Because the policy network in RC-Explainer is shared across all graph instances, while GNNExplainer needs to retrain the attention network for each instance and PGM-Explainer requires a large number of random perturbations to generate explanations;
(2) On REDDIT-MULTI-5K, RC-Explainer is slower on large-scale graphs due to the huge action space. We can solve this issue by pruning the action space, which is left to future work;
(3) Gradient-based methods generate much faster than most parametric methods.

\subsubsection{\textbf{Evaluation \wrt Visual Inspection}}\label{sec:qualitative-analyses}
\begin{figure}[t]
    \centering
	\includegraphics[width=0.8\columnwidth]{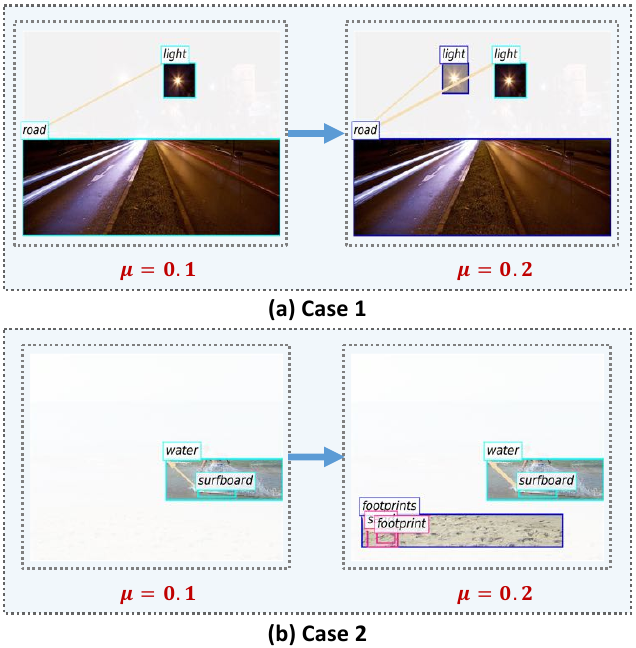}
	\vspace{-5pt}
	\caption{\wx{Showcasing the failure cases of RC-Explainer.}}
	\label{fig:failure-cases}
	\vspace{-15pt}
\end{figure}

In this section, we conduct the visual inspections of two examples in Mutagenicity and Visual Genome to give an intuitive impression of explanations.
As demonstrated in Figure~\ref{fig:case-study}, we find that:
\begin{itemize}
    \item RC-Explainer is able to capture the edges that plausibly cause the GNN's prediction. For example, when interpreting the prediction of \emph{Street} in Visual Genome, it assigns the most convincing edges (\emph{car, on, road}) and (\emph{van, in, street}) with the largest attribution scores. Whereas, the other methods easily distribute attention on \emph{tree}- or \emph{human}-related edges, as \emph{tree} and \emph{human} frequently occur with \emph{street}. However, using such spurious correlations as causation fails to interpret the prediction reliably. This again verifies the importance of causal interpretability.
    
    \item The sequential decision process enables RC-Explainer to measure the relationships of candidate edges and previous selections, thus identifying the coalition of edges and avoiding the redundant edges. For instance, when explaining the prediction of \emph{mutagenic} in Mutagenicity, RC-Explainer can highlight the functional groups, such as two \emph{NO$_2$} chemical groups, responsible for the mutagenic property~\cite{debnath1991structure}. Whereas, at most one \emph{NO$_2$} is captured by the baselines, as they hardly consider the coalition effect of an edge combination. As for the non-mutagenic molecules, RC-Explainer tends to pick bonds like carbon-hydrogen (\emph{C-H}). This makes sense because the GNN models will not predict the subgraphs with these chemical bonds as mutagenic.
    
    \item Moreover, we probe the sequential decision process of generating explanations of \emph{mutagenic}. Figure \ref{fig:sequence-show} illustrates the process, with step-wise rewards and probabilities of edges being added. Clearly, at the first two steps, RC-Explainer selects the \emph{N=O} bonds from two groups; afterward, it completes the \emph{NO$_2$} groups by adding the other \emph{N=O} bonds. Compared to the influential \emph{N=O} bonds, others like \emph{C-H} contribute less to the prediction.
    
    \item \wx{We also present two failure cases of RC-Explainer in Figure \ref{fig:failure-cases}, to show the potential out-of-distribution issue.
    In these cases of Visual Genome, RC-Explainer first picks up the most convincing edges (\emph{light, on, road}) and (\emph{surfboard, on, water}) when $\mu=0.1$, which can almost recover the target predictions of \emph{Street} and \emph{Surfing}.
    It then chooses (\emph{light, on, road}) and (\emph{footprint, on, sand}) when $\mu=0.2$, since these edges frequently occur with the target scenes.
    However, the later edges are not only redundant or spurious to introduce noises into the explanations, but also disjoint with the previous selection to make the subgraphs out of the original distribution.
    Such cases plausibly cause the fluctuations in the accuracy, especially the steep drop between $\mu=0.1$ and $\mu=0.2$ in Figure \ref{fig:acc-curve}.

    }
\end{itemize}

\section{Conclusion}

In this work, we focused on explaining the predictions made by graph neural networks.
We framed the explanation as a sequential decision process and proposed a novel reinforcement learning agent, RC-Explainer.
\wx{It approaches better explanations of GNN predictions by considering the causal effect of edges and dependencies among edges. As such, the policy network of RC-Explainer learns the causal screening strategy to efficiently yield the inﬂuential sequence of edges.
We offered extensive experiments to evaluate the explanations from different dimensions, such as predictive accuracy, contrastivity, sanity check, and visual inspections.}

Although doing interventions is effective to evaluate the causal effects of graph structures, the intervened substructures might be outside of the training distribution and cause the out-of-distribution issue \cite{DIR}.
In the future, we will exploit counterfactual generation and reasoning to solve this issue.
Moreover, we will explore the feature coalitions and interactions from the game-theoretic perspective.



\vspace{-10pt}
\ifCLASSOPTIONcompsoc
  \section*{Acknowledgments}
\else
  \section*{Acknowledgment}
\fi

This work is supported by the National Key Research and Development Program of China (2020AAA0106000), the National Natural Science Foundation of China (U19A2079, U21B2026).

\vspace{-10pt}

\bibliographystyle{IEEEtran}
\bibliography{ms}




%

\vspace{-50pt}

\begin{IEEEbiography}[{\includegraphics[width=1in,height=1.25in,clip,keepaspectratio]{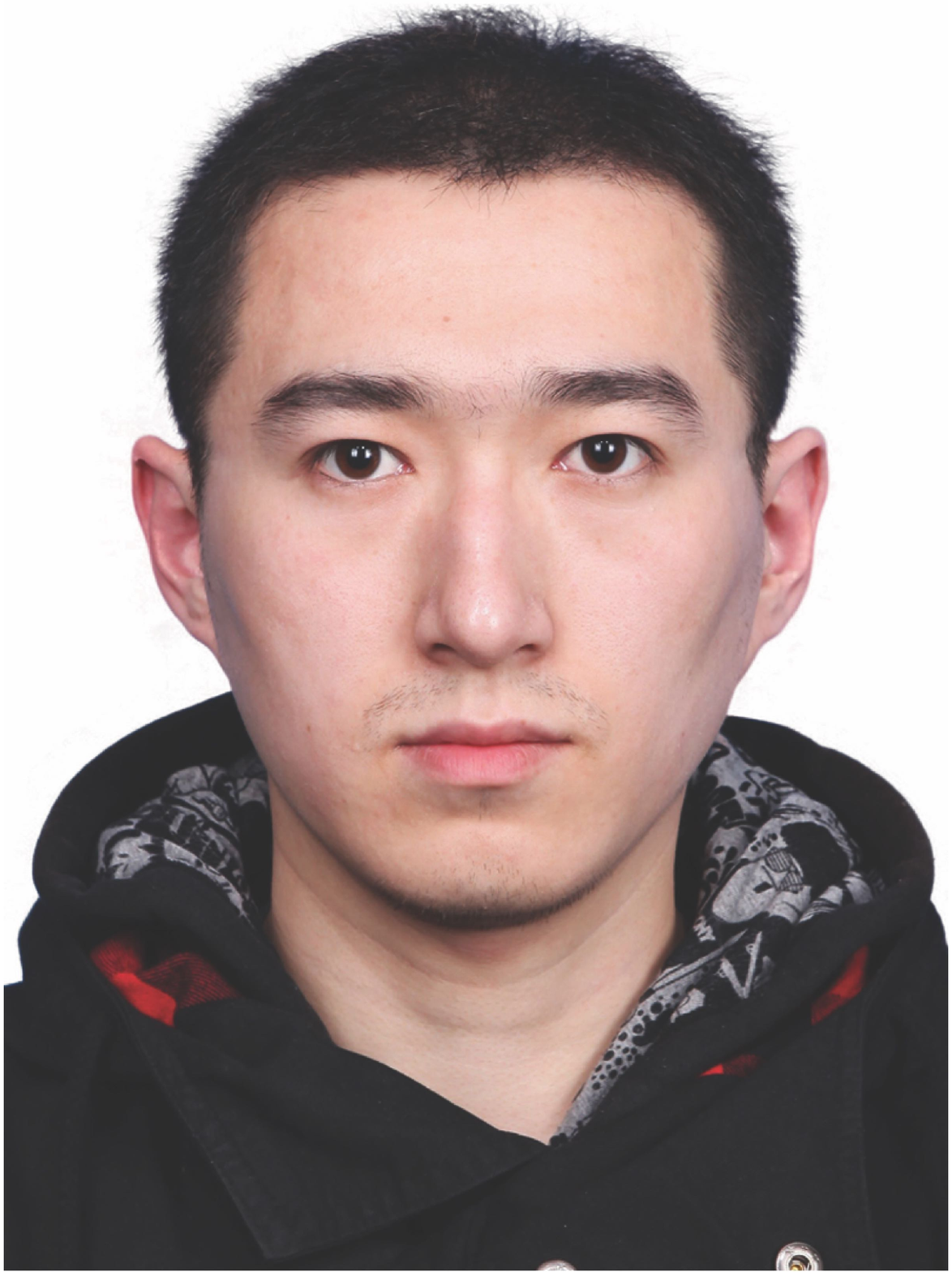}}]{Xiang Wang}
    is now a professor at the University of Science and Technology of China (USTC). He received his Ph.D. degree from National University of Singapore in 2019. His research interests include recommender systems, graph learning, and explainable deep learning techniques. He has published some academic papers on international conferences such as NeurIPS, ICLR, KDD, WWW, SIGIR. He serves as a program committee member for several top conferences such as KDD, SIGIR, WWW, and IJCAI, and invited reviewer for prestigious journals such as TKDE, TOIS, TNNLS.
\end{IEEEbiography}
\vspace{-30pt}


\begin{IEEEbiography}[{\includegraphics[width=1in,height=1.25in,clip,keepaspectratio]{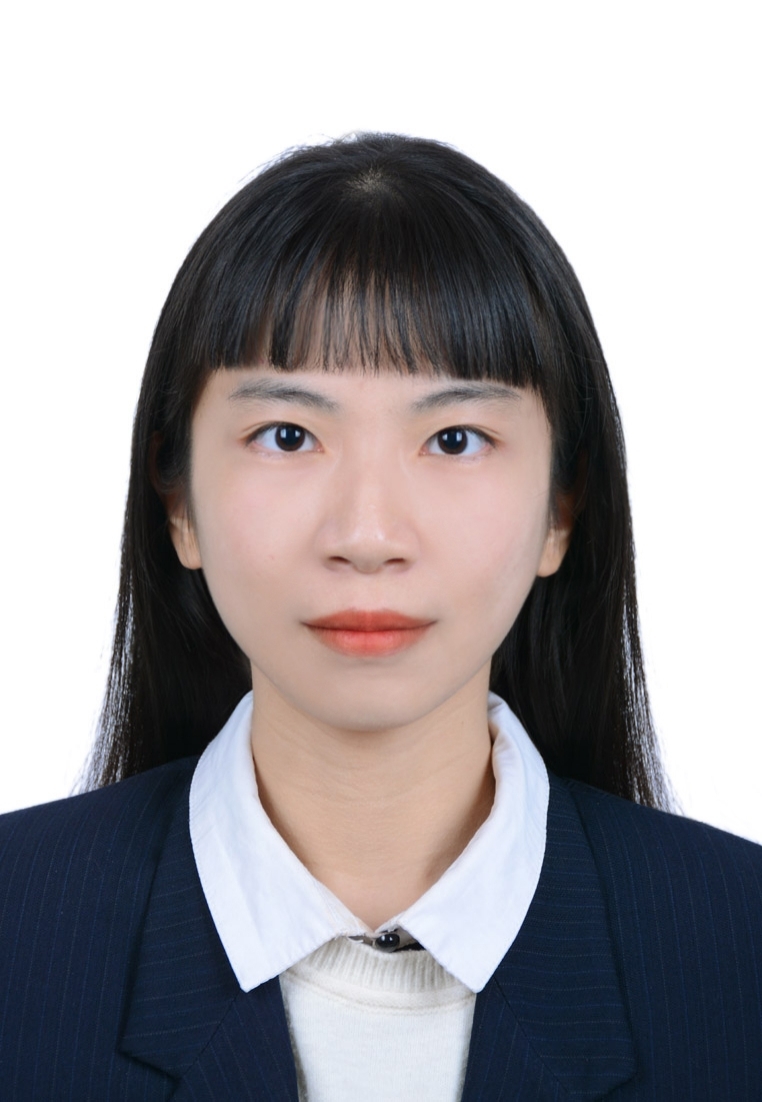}}]{Yingxin Wu} is currently a senior student at School of Data Science, University of Science and Technology of China (USTC). Her research interests includes Causal Inference and Explainable AI techniques. She has been awarded the 2020 Chen Linyi Scholarship and 2021 China National Scholarship.
\end{IEEEbiography}
\vspace{-30pt}

\begin{IEEEbiography}[{\includegraphics[width=1in,height=1.25in,clip,keepaspectratio]{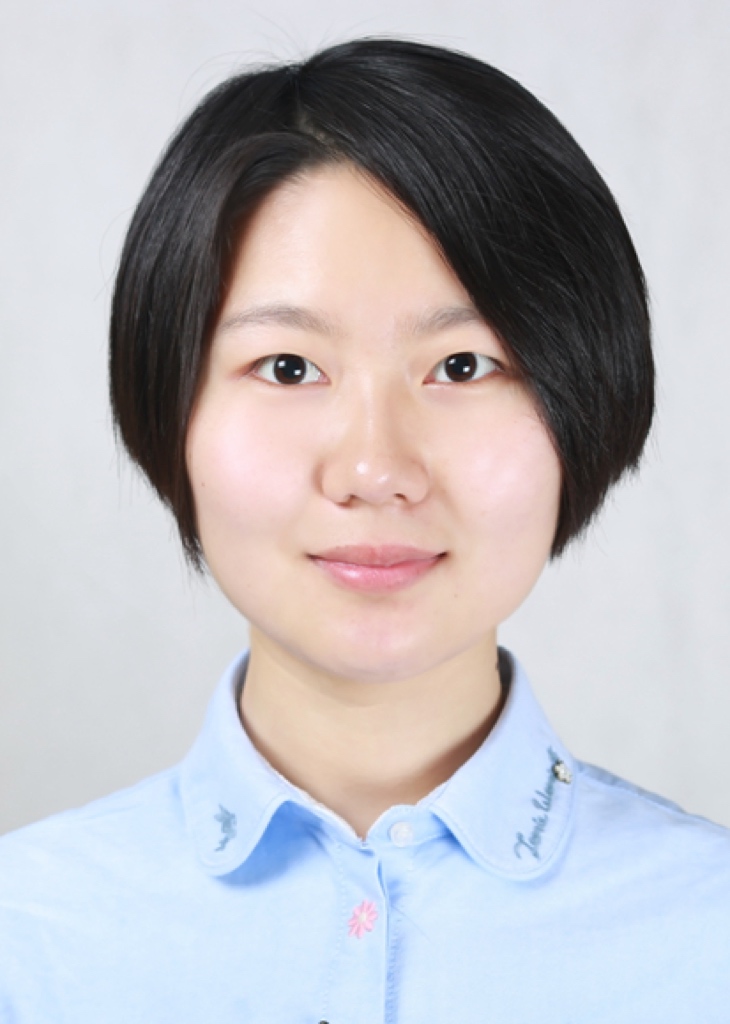}}]{An Zhang}
    is now a research fellow at National University of Singapore.
    She received her Ph.D. degree from the Department of Statistics and Applied Probability, National University of Singapore, in 2021.
    Her research interests include explainable artificial intelligence, security, causal effect, and graph neural network.
    She has publications appeared in several top conferences such as NeurIPS, ICLR, SIGIR.
\end{IEEEbiography}
\vspace{-30pt}

\begin{IEEEbiography}[{\includegraphics[width=1in,height=1.25in,clip]{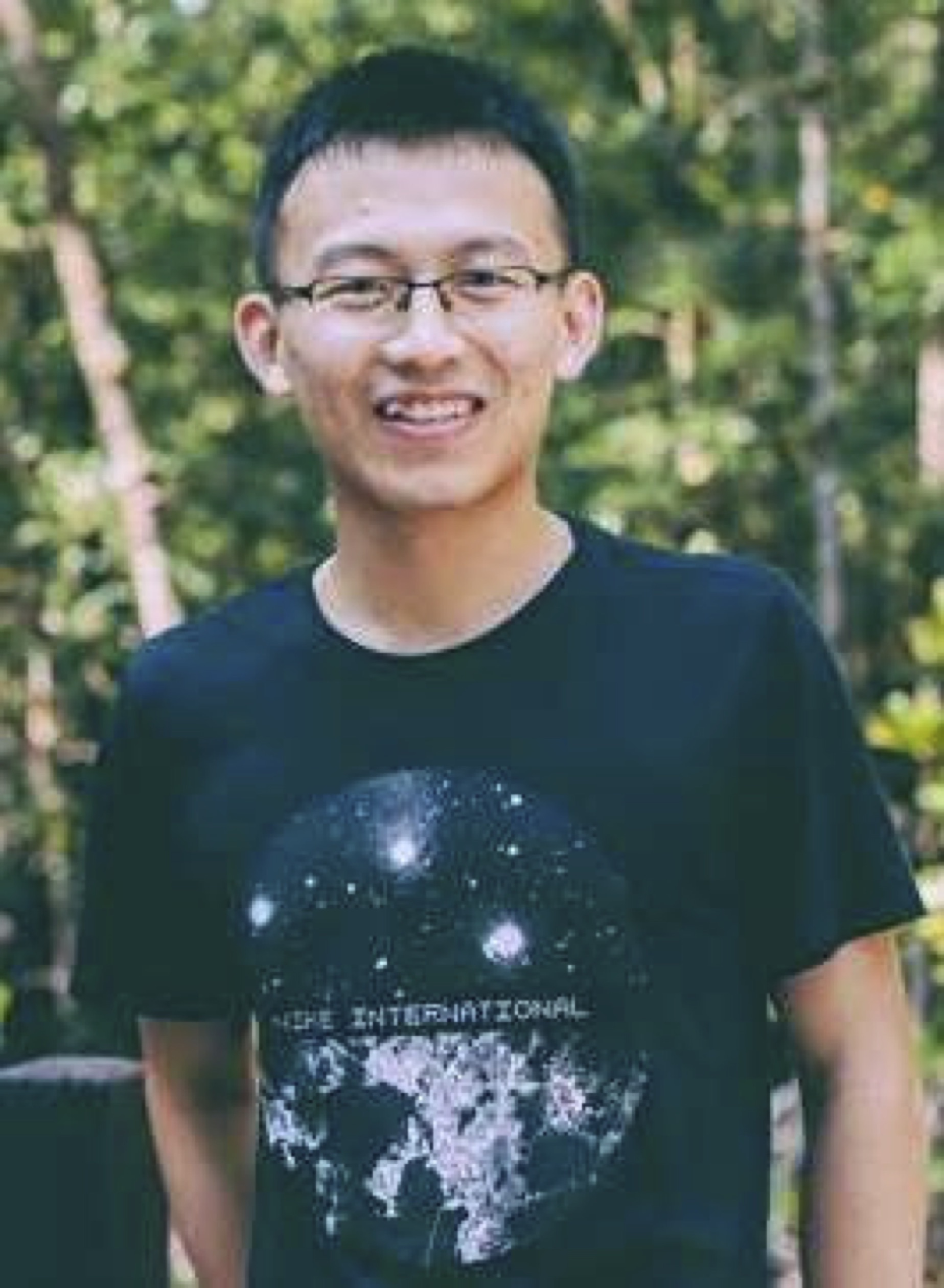}}]{Fuli Feng}
    is a professor at the University of Science and Technology of China (USTC). He received Ph.D. in Computer Science from NUS in 2019. His research interests include information retrieval, data mining, and multi-media processing. He has over 30 publications appeared in several top conferences such as SIGIR, WWW, and MM, and journals including TKDE and TOIS. His work on Bayesian Personalized Ranking has received the Best Poster Award of WWW 2018. Moreover, he has been served as the PC member for several top conferences including SIGIR, WWW, WSDM, NeurIPS, AAAI, ACL, MM, and invited reviewer for prestigious journals such as TOIS, TKDE, TNNLS, TPAMI, and TMM.
\end{IEEEbiography}
\vspace{-30pt}

\begin{IEEEbiography}[{\includegraphics[width=1in,height=1.25in,clip,keepaspectratio]{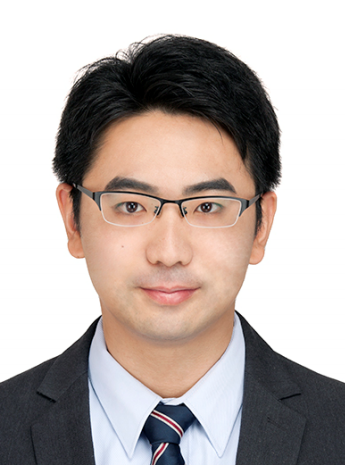}}]{Xiangnan He}
    is a professor at the University of Science and Technology of China (USTC). His research interests span information retrieval, data mining, and multi-media analytics. He has over 90 publications in top conferences such as SIGIR, WWW, and MM, KDD, and journals including TKDE, TOIS, and TMM. His work has received the Best Paper Award Honorable Mention in WWW 2018 and SIGIR 2016. He is in the Editorial Board of the AI Open journal, served as the PC chair of CCIS 2019, the area chair of MM 2019, ECML-PKDD 2020, and the (senior) PC member for top conferences including SIGIR, WWW, KDD, WSDM etc.
\end{IEEEbiography}
\vspace{-30pt}

\begin{IEEEbiography}[{\includegraphics[width=1in,height=1.25in,clip,keepaspectratio]{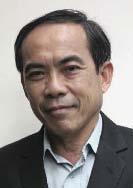}}]{Tat-Seng Chua} is the KITHCT Chair Professor at the School of Computing, National University of Singapore. He was the Acting and Founding Dean of the School during 1998-2000. Dr Chuas main research interest is in multimedia information retrieval and social media analytics. In particular, his research focuses on the extraction, retrieval and question-answering (QA) of text and rich media arising from the Web and multiple social networks. He is the co-Director of NExT, a joint Center between NUS and Tsinghua University to develop technologies for live social media search. Dr Chua is the 2015 winner of the prestigious ACM SIGMM award for Outstanding Technical Contributions to Multimedia Computing, Communications and Applications. He is the Chair of steering committee of ACM International Conference on Multimedia Retrieval (ICMR) and Multimedia Modeling (MMM) conference series. Dr Chua is also the General Co-Chair of ACM Multimedia 2005, ACM ICMR 2005, ACM SIGIR 2008, and ACM Web Science 2015. He serves in the editorial boards of four international journals. Dr. Chua is the co-Founder of two technology startup companies in Singapore. He holds a PhD from the University of Leeds, UK.
\end{IEEEbiography}
    
    
    

\appendices

\begin{table*}[t]
  \centering
  \caption{Model architectures of target models and RC-explainer}
  \label{tab:model-reproducibility}
  \resizebox{0.8\textwidth}{!}{
  \begin{tabular}{c|l|c|c|c}
  \toprule
  &Datasets & Mutag & BA3-motif & Reddit-5k\\
  \midrule
  \multirow{2}{*}{Target Model} & Type of GNN & PTGNN \cite{DBLP:conf/iclr/HuLGZLPL20} &  ASAP \cite{ASAP} & GCN \cite{GCN} \\
  & \#Neurons of GNN & [75,32,75,32,16,2] & [64,64,64,128,3] & [32,32,32,64,5]\\
  \midrule
  \multirow{4}{*}{RC-Explainer} & Type of GNN $g$ & PTGNN \cite{DBLP:conf/iclr/HuLGZLPL20} &  ASAP \cite{ASAP} & GCN \cite{GCN} \\
  & \#Neurons of GNN $g$ & [75,32,75,32,16,2] & [64,64,64,128,3] & [32,32,32,64,5]\\
  & \#Neurons of MLP$_1$ & [64,128,64,32] &  [128,256,128,64] & [64,128,64,32]\\
  & \#Neurons of MLP$_{2,c}$ & [32,32,2] & [64,64,3]& [32,32,5]\\
  \bottomrule
  \end{tabular}}
\end{table*}

\begin{table*}[t]
  \centering
  \caption{Hyperparameter settings of RC-explainer}
  \label{tab:hyperparameter-reproducibility}
  \resizebox{0.98\textwidth}{!}{
  \begin{tabular}{c|l|c|c|c}
  \toprule
  &Datasets & Mutag & BA3-motif & Reddit-5k\\\midrule
  \multirow{4}{*}{RC-Explainer} & Learning Rate $lr$ & $\{0.00001,\textbf{0.0001},0.001\}$ & $\{0.00001,0.0001,\textbf{0.001}\}$ & $\{0.00001,\Mat{0.0001},0.001\}$ \\
  & Weight Decay $l_2$ & $\{\textbf{0.00001},0.0001,0.001\}$ & $\{0.00001,\textbf{0.0001},0.001\}$ & $\{0.00001,\Mat{0.0001},0.001\}$\\
  & Reward Mode & \{\textbf{MI}, Binary, CE\} & \{\textbf{MI}, Binary, CE\} & \{\textbf{MI}, Binary, CE\}\\
  & Beam Search & $\{2,4,\Mat{8},16\}$ & $\{2,4,\Mat{8},16\}$ & $\{2,4,\Mat{8},16\}$\\
  \bottomrule
  \end{tabular}}
\end{table*}

\section{Reproducibility}
We have released our codes and datasets at \url{https://github.com/xiangwang1223/reinforced_causal_explainer} to ensure the reproducibility.

We summarize the model architectures in Table \ref{tab:model-reproducibility}, where the target model is being explained via RC-Explainer.
Within RC-Explainer, $g$ is the GNN model to perform the representation learning, MLP$_1$ is the multilayer perceptron to generate the representations for action candidates, while MLP$_{2,c}$ is the class-specific multilayer perceptron to yield the importance score of each action candidate.

Moreover, we present the hyperparameter settings of RC-Explainer in Table \ref{tab:hyperparameter-reproducibility}, where $\{\cdot\}$ indicates the range of tuning each hyperparameter and the \textbf{bold} numbers are our final settings.



\vspace{-10pt}

\end{document}